\documentclass[twoside,11pt]{article}

\usepackage{blindtext}

\usepackage{jmlr2e}

\usepackage{xcolor}

\usepackage{soul} 

\usepackage{amsfonts}
\usepackage{dsfont}
\usepackage{float}
\usepackage{subcaption}
\usepackage{algorithm}
\usepackage{algorithmic}

\usepackage{amsmath}
\DeclareMathOperator*{\argmax}{arg\,max}

\usepackage{amsfonts}
\usepackage{dsfont}
\usepackage{float}
\usepackage{subcaption}
\usepackage{algorithm}
\usepackage{algorithmic}

\usepackage{amsmath}
\usepackage{pgffor}

\newtheorem{theorem}{Theorem}
\newtheorem{corollary}{Corollary}
\newtheorem{lemma}{Lemma}
\newtheorem{remark}{Remark}
\newtheorem{definition}{Definition}

\newtheorem{assumption}{Assumption}

\usepackage{lastpage}
\jmlrheading{23}{2022}{1-\pageref{LastPage}}{1/21; Revised 5/22}{9/22}{21-0000}{Author One and Author Two}

\ShortHeadings{Is Prior-Free Black-Box NS-RL Feasible?}{Gerogiannis, Huang, Veeravalli}
\firstpageno{1}

\begin{document}

\title{Is Prior-Free Black-Box Non-Stationary Reinforcement Learning Feasible?}

\author{\name Argyrios Gerogiannis \email ag91@illinois.edu \\
       \addr Department of Electrical and Computer Engineering\\
        University of Illinois Urbana-Champaign\\
        Champaign, IL 61820, USA
       \AND
       \name Yu-Han Huang \email yuhanhh2@illinois.edu \\
       \addr Department of Electrical and Computer Engineering\\
        University of Illinois Urbana-Champaign\\
        Champaign, IL 61820, USA
        \AND
        \name Venugopal V. Veeravalli \email vvv@illinois.edu \\
       \addr Department of Electrical and Computer Engineering\\
        University of Illinois Urbana-Champaign\\
        Champaign, IL 61820, USA
        \AND}

\def\fundingtext{This work  was supported by the Army Research Laboratory under Cooperative Agreement W911NF-17-2-0196, and by a grant from the C3.ai Digital Transformation Institute, through the University of Illinois at Urbana-Champaign.}

\maketitle

\begin{abstract}
We study the problem of Non-Stationary Reinforcement Learning (NS-RL) without prior knowledge about the system’s non-stationarity. A state-of-the-art, black-box algorithm, known as MASTER, is considered, with a focus on identifying the conditions under which it can achieve its stated goals. Specifically, we prove that MASTER's non-stationarity detection mechanism is not triggered for practical choices of horizon, leading to performance akin to a random restarting algorithm. Moreover, we show that the regret bound for MASTER, while being order optimal, stays above the worst-case linear regret until unreasonably large values of the horizon. To validate these observations, MASTER is tested for the special case of piecewise stationary multi-armed bandits, along with methods that employ random restarting, and others that use quickest change detection to restart. A simple, order optimal random restarting algorithm, that has prior knowledge of the non-stationarity is proposed as a baseline. The behavior of the MASTER algorithm is validated in simulations, and it is shown that methods employing quickest change detection are more robust and consistently outperform MASTER and other random restarting approaches.
\end{abstract}


\section{INTRODUCTION}
In the Reinforcement Learning (RL) problem an agent interacts with an environment with the goal of maximizing some notion of cumulative reward \citep{sutton2018reinforcement}. Standard RL algorithms assume that the environment is stationary in time. However, the stationarity assumption may not be accurate in many practical settings, and therefore the non-stationary version of the RL problem has received some attention in recent years.

The problem of Non-Stationary RL (NS-RL) has been studied across various settings, and different solutions have been proposed \citep{auer2019adswitch,chen2019context,yadkori2023nsb,cheung2019ns,russac2020genLbs,faury2021glbs,besson2022efficient,mao21qrucb,chen2022goalorient,touati2020LMDPs,cheung2020optimism,cheng2023lowrank,wei2021non}. A key factor distinguishing solution methods is their approach to mitigating the underlying non-stationarity. Based on this characteristic, methods can be broadly classified into \emph{adaptive} approaches and \emph{restarting} approaches. The former involve designing algorithms that are able to run continuously without restarting, while adjusting their performance, whereas the latter employ a mechanism according to which the algorithm restarts the learning process at certain points in time.

In the recent work of \citet{peng_papadim_2024}, it has been shown that restarting the learning process could be better than continuously adapting the learning, as the complexity of restarting based procedures is often comparable to or better than adaptive procedures in the worst-case scenario. Due to this observation, the focus of our work is on restarting based approaches. Under this paradigm, a critical task of the NS-RL problem is the development of an appropriate restarting mechanism for a given setting. The design of the decision rule that leads to the restart is intricate. A well-designed restarting algorithm should reset after the non-stationarity has affected the environment significantly, while incurring as small a delay as possible. To decide when to restart, the algorithm can either have prior information about the non-stationarity or be \emph{prior-free}, i.e., it operates without any prior knowledge of the non-stationarity. The prior-free approach is more applicable in practice, and there have been a number of recent works on providing order optimal, prior-free solutions in different NS-RL settings \citep{wei2021non,chen2019context,auer2019adswitch,yadkori2023nsb,besson2022efficient}. 

In these works on prior-free solutions, various non-stationarity detection tests have been proposed \citep{auer2019adswitch,yadkori2023nsb,wei2021non}. Works in Non-Stationary Multi-Armed Bandits (NS-MABs) have employed Quickest Change Detection (QCD) \citep{poor-hadj-qcd-book-2009,veeravalli2014qcd} techniques to restart in piecewise stationary settings. In fact, the QCD-based approach of \citet{besson2022efficient} in NS-MABs under piecewise stationarity, is the state-of-the-art method, which not only significantly outperforms the influential prior-free approach proposed in \citet{auer2019adswitch} but also demonstrates better detection capabilities.

In the general NS-RL literature, the method introduced by \citet{wei2021non} stands out. This prior-free method, termed as MASTER, is applicable across various NS-RL settings and achieves order optimal performance. This is the only known method with such generic black-box capabilities: as long as the input algorithm satisfies MASTER's assumptions, it can be employed in a seamless manner. MASTER was applied to various RL settings\footnote{MASTER was also applied to inifinite-horizon Markov Decision Processes, but it is not order optimal under prior-free assumptions. Thus, its analysis under this setting will not be studied in this work.} including: Multi-armed Bandits \citep{auer2002finite}, Linear Bandits \citep{yadkori2011improved}, Generalized Linear Bandits (LBs) \citep{filippi2010parametric}, Finite-horizon Tabular Markov Decision Processes \citep{jin2018qlearning}, Finite-horizon Linear Markov Decision Processes (LFMDPs) \citep{jin2020provably} and Contextual Bandits \citep{agarwal2014taming,simchi2022bypassing}. The paper by \citet{wei2021non} lacks experimental validation. A few works have tested MASTER's practical effectiveness, under the settings of LBs \citep{pmlr-v206-wang23k} and LFMDPs \citep{zhou2022nonstationary}. The absence of empirical validation in other settings motivates the need to assess MASTER's practical feasibility.

Our goal in this paper is to evaluate the performance bounds for MASTER given in \citet{wei2021non}, and assess its practical effectiveness through experimental validation. 
In summary, the primary contributions of this work are:
\begin{itemize}
    \item We prove that MASTER's non-stationarity detection mechanism is not triggered for practical choices of horizon, leading to performance akin to a random restarting algorithm. 
    \item We show that the regret bound for MASTER given in \citet{wei2021non}, while being order optimal, stays above the worst-case linear regret until unreasonably large values of the horizon.
    \item We run experiments for the special NS-RL problem of Piecewise Stationary MABs (PS-MABs), and validate the performance issues of MASTER.
    \item We propose an order optimal, non-prior-free, random restarting algorithm for PS-MABs, that can be used as a baseline for the problem.
    \item We show in simulations for PS-MABs that methods which employ QCD techniques are more robust and superior in performance compared to MASTER and the order optimal, random restarting approach.
\end{itemize}

\section{PRELIMINARIES}
\label{sec:prelim}
\noindent \textbf{Notation}: For any non-negative integers $n,e$, $[n]$ denotes the set $\{1,2,\dots,n\}$ and $[n,e]$ corresponds to $\{n,n+1,\dots,e\}$. The indicator function is denoted by $\mathds{1}{[\cdot]}$. We write $h_1(x)=\Tilde{\mathcal{O}}(h_2(x))$ with high probability, if $h_1(x)=\mathcal{O}(\mathrm{poly}(\log(T/\delta))h_2(x))$ with probability $1-\delta$ for some $\delta\in (0,1)$.
\subsection{Problem Formulation And Definitions}
We formally define the NS-RL problem as follows.
\begin{definition} Assume an agent that interacts with an environment through a behavior-policy set $\Pi$ over a horizon $T$, that is, a fixed number of time steps. The environment has a collection of $T$ reward functions $f_1,...,f_T:\Pi \rightarrow [0,1]$, unknown to the agent, that model the environment's feedback. In each time step $t=1,...,T$, the agent selects a policy $\pi_t\in \Pi$ and receives the feedback in the form of a random reward $R_t\in [0,1]$ with mean $f_t(\pi_t)=\mathbb{E}[R_t|\pi_t]$. The goal is to minimize the dynamic regret,
    $\mathrm{Reg}_{\mathrm{D}}(T)=\mathbb{E}\left[\sum_{t=1}^T (f_t^*-f_t(\pi_t))\right]$, where $f_t^*=\max_{\pi\in\Pi} f_t(\pi)$ is the best expected reward at time $t$.
    \label{def:ns_env}
\end{definition}
To measure the underlying non-stationarity, a general measure is provided as follows.
\begin{definition}
   (Following \citet{wei2021non}.) $\Delta:[T]\rightarrow \mathbb{R}$ is a non-stationarity measure, if it satisfies $\Delta(t)\geq \max_{\pi \in \Pi} |f_t(\pi)-f_{t+1}(\pi)|$,  for all $t$. For interval $[s,e]$, the cumulative non-stationarity is given as $\Delta_{[s,e]}=\sum_{\tau=s}^{e-1} \Delta(\tau)$, with $\Delta_{[s,s]}=0$, and the number of changes ($+$1) as $L_{[s,e]}=1+\sum_{\tau=s}^{e-1}\mathds{1}{[\Delta(\tau)\neq 0]}$.
    \label{def:ns_measure}
\end{definition}

For a given horizon $T$, the total cumulative non-stationarity and the number of changes are denoted as $\Delta=\Delta_{[1,T]}$ and $L=L_{[1,T]}$, respectively. In every setting except infinite-horizon MDPs, the dynamic regret minimax bound in $\tilde{\mathcal{O}}(\cdot)$ is given by 
$\min\{\mathrm{Reg}^*_L,\mathrm{Reg}^*_\Delta \}$, where 
$\mathrm{Reg}^*_L=\sqrt{LT}$ and 
$\mathrm{Reg}^*_\Delta=\Delta^{1/3}T^{2/3}+\sqrt{T}$.  These bounds are known to be optimal even when $L$ and $\Delta$ are known \citep{wei2021non}. Thus, the goal in prior-free NS-RL is to achieve the optimal bound when $L$ and $\Delta$ are unknown, i.e., an algorithm that is both prior-free and order optimal.

\subsection{The MASTER Algorithm}

\label{subsec:MASTER}
MASTER is a black-box reduction that schedules and runs multiple instances of an input algorithm, denoted as ALG, and uses two tests for non-stationarity detection. More specifically, MASTER is scheduled for blocks of length $2^n$ in the horizon, where $n\in \mathbb{N}$. The scheduling of multiple instances is done by first partitioning the interval of length $2^n$ equally into $2^{n-m}$ sub-intervals for every $m=0,\dots,n$. Then, for each sub-interval of length $2^{m}$, an instance of ALG with length $2^m$ is scheduled with probability $\frac{\rho (2^{n})}{\rho (2^{m})} \leq 1$ for that specific sub-interval. Since $\frac{\rho(2^n)}{\rho(2^n)}=1$, there always exists a length-$2^n$ instance, and, depending on the random scheduling, possibly multiple length-$2^m$ instances.

After the scheduling and the initialization, the instance with the shortest length that covers the current time step is made active, while all the other instances remain inactive. MASTER follows the policy of the active instance and updates it accordingly, whereas all inactive instances are paused. Finally, MASTER performs the two non-stationarity tests, termed as Test 1 and Test 2, at the end of each time step and if either of their conditions is satisfied, MASTER restarts the described procedure with $n=n+1$. We provide the analytic algorithm of MASTER in the Appendix.

MASTER, as a black-box method, has specific requirements for ALG, presented bellow.

\begin{assumption}\label{assumption_one}
(Following \citet{wei2021non}.) ALG outputs $\tilde{f}_t\in [0,1]$ at the beginning of each time step $t$. Suppose there exists a non-increasing function $\rho: [T] \rightarrow \mathbb{R}$ such that $\rho(t) \geq 1/\sqrt{t}$ and $C(t) = t \rho(t)$ is non-decreasing. Additionally, assume $\Delta(t)$ to be a non-stationary measure. Then, for any $t \in [T]$ such that $\Delta_{[1,t]} \leq \rho(t)$, the following holds with probability at least $1-\delta/T$:
\begin{enumerate}
    \item $\tilde{f}_t \geq \min _{\tau\in[1,t]}f^*_\tau - \Delta_{[1,t]}$, and
    \item $\frac{1}{t}\sum_{\tau=1}^t\left( \tilde{f}_\tau - R_\tau \right)\leq \rho(t) + \Delta_{[1,t]}$.
\end{enumerate}
\end{assumption}
This assumption implies that if there exists an interval where the non-stationarity is weak, ALG should be able to achieve an appropriate regret bound and provide an optimistic estimate of the optimal reward. Since the active instance of ALG at each time step may be different, the notation $\tilde{g}_t$ is used to denote the scalar output $\tilde{f}_t$ of the current active instance of MASTER. MASTER's approach is able to ensure an equivalence of the guarantees of Assumption 1 through a procedure termed as MALG. MALG is essentially the random scheduling scheme, i.e., MASTER without the non-stationarity detection tests. MALG is designed in such a way that its performance is equivalent to a multi-scale version of the Assumption 1, as shown in Lemma 3 from \citet{wei2021non} which we present below.

\begin{lemma}(Adapted from \citet{wei2021non}.)
Let \(\hat{n} = \log_2 T + 1\), \(\hat{\rho}(t) = 6\hat{n} \log(T/\delta)\rho(t)\) and \(t' = t - alg.s + 1\). Suppose Assumption \ref{assumption_one} holds for ALG, and that \(n \leq \log_2 T\). Then, for any instance $alg$, with start at $alg.s$ and finish at $alg.e$, that is maintained by MALG and any \(t \in [alg.s, alg.e]\) that satisfies $\Delta_{[alg.s,t]} \leq \rho(t')$, MALG satisfies with with probability at least $1-\delta/T$:
\begin{enumerate}
    \item $\tilde{g}_t \geq \min_{\tau \in [alg.s,t]} f^*_\tau - \Delta_{[alg.s,t]}$, and
    \item $\frac{1}{t'} \sum_{\tau = alg.s}^t (\tilde{g}_\tau - R_\tau) \leq \hat{\rho}(t') + \hat{n} \Delta_{[alg.s,t]}$.
\end{enumerate}
\label{lemma_3_master}
\end{lemma}
Thus, it is evident that for an instance $alg$ of MASTER, if there exists $t\in [alg.s,alg.e]$ such that $\Delta_{[alg.s,t]} \leq \rho(t')$, then that instance can ensure the appropriate regret for the interval $[alg.s,t]$. However, due to the existence of non-stationarity, this condition may be violated, and therefore MASTER needs to restart. In a prior-free setting, the exact identification of this condition is not possible, and thus MASTER's detection scheme uses the consequences of Lemma \ref{lemma_3_master} to identify the the violation of the condition.

Also, there is an assumption about the parameter $\delta$.
\begin{assumption}\label{assum_two}
    (Following \citet{wei2021non}.) The probability parameter $\delta \in (0,1)$ is assumed to have an order of $1/\mathrm{poly}(T)$.
\end{assumption}

\section{THEORETICAL ANALYSIS}
\label{sec:theory}

\subsection{MASTER's Non-Stationarity Detection}
\label{sec:ns_detect}
Regarding MASTER's analysis, it is necessary to ensure its non-stationarity detection scheme works effectively. MASTER employs two detection tests. The first, Test 1, leads to a restart when for any $2^m$-length instance $alg$, both $t = alg.e$ and the following hold:
\begin{equation*}
    \frac{1}{2^m}\sum_{\tau=alg.s}^{alg.e}R_\tau -\min_{\tau\in[t_n,t]} \tilde{g}_{\tau}\geq 9\hat{\rho}(2^m).
\end{equation*}
The second test, Test 2, is applied at each time step $t$ of block $2^n$, with the following restarting condition:
\begin{equation*}
    \frac{1}{t-t_n+1}\sum_{\tau=t_n}^t (\tilde{g}_\tau -R_\tau)\geq 3\hat{\rho}(t-t_n+1)
\end{equation*}
where $t_n$ is the start of the current $2^n$ block and $t\in[t_n,t_n+2^n-1]$.

To detect significant non-stationarity using these tests, the statistics on the LHS need to surpass the threshold on the RHS. However, due to MASTER's assumptions, the LHS has an upper bound. Since $\rho(t)$ is non-increasing and $\rho(t) \geq 1/\sqrt{t}$, this enforces a constraint on $\delta$, irrespective of the form of $\rho(t)$. On the other hand, fixing a specific order dependence for $\delta$ sets a bound on $T$ for which the tests can be triggered.

\begin{theorem}\label{thrm:theor_1}
    If Assumption \ref{assumption_one} is satisfied, the thresholds of Test 1 and Test 2 can be crossed only if $\delta\geq T \exp\left(- \frac{\sqrt{T}}{54(\log_2T+1)} \right)$, and since $\delta<1$, $T$ must be at least 1.24 billion.
\end{theorem}
The proof of this theorem is given in the Appendix. To prove this theorem, we also present the following auxiliary lemma, which provides additional insights into the non-stationarity detection tests.

\begin{lemma}\label{lemma:test_work}
    Test 1 will not be triggered for any value of the horizon that satisfies $\rho(T)>\frac{1}{54(\log_2T +1)\log(T/\delta)}$ and Test 2 will also not be triggered if $\rho(T)>\frac{1}{18(\log_2T +1)\log(T/\delta)}$.
\end{lemma}

Theorem \ref{thrm:theor_1} sets additional dependency for $\delta$ with $T$ on top of Assumption \ref{assum_two}, and it shows that the non-stationary detection scheme cannot work for a significant range of possible horizons, highlighting possible practical issues. These results are derived assuming that the statistics of the tests on the LHS of the test inequalities are maximized and the threshold on the RHS is minimized, to identify the best conditions under which the thresholds are crossed. Fixing a dependency of $1/\mathrm{poly}(T)$ for $\delta$, can lead to a higher lower bound for the horizon, as can be seen in the inequality of Theorem \ref{thrm:theor_1}. In addition, since the range of the statistics used for detection may be lower than when the LHS of the test inequalities is maximized, the horizon can be much more constrained. For example, if a case where the statistics of the tests range in $[0,0.1]$ the minimum horizon value to detect the non-stationarity would be around at least 313 billion.
\begin{remark}\label{rem:remark_one}
    The lower bound on $T$ from Theorem \ref{thrm:theor_1}  may be even larger as it depends on the form of the non-increasing function $\rho(t)$ and the dependence of $\delta$ with $T$. For instance, using the proposed functions in \citet{wei2021non} for MABs, in the simplest case of a two-armed bandit, $T$ needs to be at least 143 billion.
\end{remark}

Regardless of horizon length, the two non-stationarity detection tests also impose two different conditions on the size of the main $2^n$-length instance as well as the length of the $2^m$-length instances. Thus, we have the following corollary.
\begin{corollary}\label{cor:no_detect}
     Any $2^m$-length instance, such that $\rho(2^m)> \frac{1}{54(\log_2T +1)\log(T/\delta)}$, will not trigger Test 1, and any $2^n$-block such that $\rho(2^n)> \frac{1}{18(\log_2T +1)\log(T/\delta)}$, will not trigger Test 2 as well.
\end{corollary}
The proof of this corollary follows immediately from the proof of Lemma \ref{lemma:test_work}. Since $\rho(t)$ is non-increasing, this Corollary sets a lower bound on the value of $n$, such that the non-stationarity detection tests can be triggered. Selecting an $n$ that is not large enough to satisfy Corollary \ref{cor:no_detect} leads to every instance not being able to identify any non-stationarity that may exist. Moreover, even if a proper $n$ is chosen, since $m$ ranges from $0$ to $n$ there will exist a set of $2^m$-instances that will not contribute in the non-stationarity detection process of Test 1 at all.

\subsection{MASTER's Performance Analysis}
\label{subsec:perform_an}
MASTER's performance is strongly tied to its non-stationary detection scheme. In particular, the regret bounds given in Theorem 2 of \citet{wei2021non} are derived through the guarantees provided by Test 1 and Test 2. 

To obtain the regret bound of MASTER, \citet{wei2021non} first derive a regret bound for any $2^n$ block by dividing the dynamic regret into two terms, and then, using Test 1 and Test 2 to bound the terms appropriately, as presented in Lemma 16 of Appendix C from \citet{wei2021non}. 

As seen in Definition \ref{def:ns_env}, the reward is bounded in $[0,1]$, and thus $T$ is a trivial upper bound for the regret over a horizon of length $T$. Therefore, any non-trivial bound on the regret should be smaller than $T$.  As we have shown in Section~\ref{sec:ns_detect}, the thresholds of the tests are highly conservative, which in turn affects the regret bounds. This is presented in the following Corollary.
\begin{corollary}\label{cor:master_perf}
    If Assumption \ref{assumption_one} holds, then the $\mathrm{Reg}_{\mathrm{D}}(T)$ bound of MASTER is at least \[
    B_\mathrm{D}(T)=24 (\log_2(T) + 1) \log (T) \sqrt{T} \left( 1 + 15 \log(T) \right).\] Therefore, whenever
    \[T\leq 4\times 10^{14},\quad T < B_\mathrm{D}(T),\] i.e., the $\mathrm{Reg}_{\mathrm{D}}(T)$ bound of MASTER is trivial.
\end{corollary}
The proof of the corollary is given in the Appendix. 
Note that similar to the statement of Remark \ref{rem:remark_one}, having a more constrained form for $\rho(t)$ will lead to a higher upper bound for Corollary \ref{cor:master_perf}. Based on this result, even though the analysis of MASTER ensures that it is order optimal, its performance may not be guaranteed for the majority of practical applications, as its performance bounds are not useful for horizons of practical interest (horizons larger than $4 \times 10^{14}$ are hardly expected in practical settings). 

MASTER may be order optimal in the $\tilde{\mathcal{O}}(\cdot)$ notation, however, it is evident that when the logarithmic terms and the constants are not omitted, the regret bound becomes trivial for an unreasonably large range of horizons. Deriving tight performance bounds is crucial, as a state-of-the-art method should achieve non-trivial regret performance for reasonable horizon values.

\subsection{The Choice Of Piecewise Stationary Multi-Armed Bandits}

To verify the performance of MASTER according to our results, it suffices to apply the algorithm to the simplest case considered in \citet{wei2021non}, which corresponds to NS-MABs. The MAB problem is a special case of the RL problem, which can considered as having only a single, known state \citep{sutton2018reinforcement}. If the performance of MASTER exhibits issues in this special case, it is reasonable to assume that the same issues occur for other NS-RL problems. 

The difficulty of the NS-MAB problem is associated with the way the non-stationarity evolves. We will consider the case of piecewise stationarity, i.e., when the changes happen at specific points and between consecutive changes the system is stationary. The piecewise stationary assumption is an important case of practical interest, as many real-world problems correspond to scenarios at which the environment remains constant between changes (see e.g. \citet{seznec2020rot,besson2022efficient}).

We should note that the choice of piecewise stationarity assists MASTER's assumptions for the existence of intervals that do not have significant non-stationarity, ensuring that $\Delta_{[alg.s, t ]} \leq \rho (t-alg.s+1)$, as shown in Lemma \ref{lemma_3_master}. 
\begin{definition}
    In a piecewise stationary environment, there exist some change-points $\{ \nu^{(k)} \in \{1, \dots, T\} :\; k \in \{1, \dots, N_{T}\} \}$, where $\nu^{(k)}$ denotes the $k^{\mathrm{th}}$ change-point and $N_{T}$ denotes the number of change-points over the finite horizon $T$, with $N_{T}=\sup\{ k\in\mathbb{N}:\nu^{( k)}\leq T\}$. Then, for any time steps between consecutive change-points, the reward functions in a piecewise stationary environment remain identical, i.e., for all $t, t'$ such that $t, t' \in \{ \nu^{(k-1)}, \dots, \nu^{(k)}-1 \}$ for some $k \in \{1, \dots, N_{T}\}$, $f_{t} = f_{t'}$.
\end{definition} 
Piecewise stationarity allows for easier modeling and adaptation compared to more complex forms of non-stationarity, where changes are continuous. In the case of gradual continuous changes for which $\Delta(t)$ remains bounded away from $0$ for all $t$, $\Delta=\mathcal{O}(T)$, and  the optimal regret will clearly be linear. Therefore, under gradual continuous changes, an NS-RL algorithm can attain sub-linear regret only if $\Delta(t)$ converges to $0$ at a sufficiently fast rate, i.e., the environment converges to some stationary state. For the piecewise stationary setting we do not require the convergence of $\Delta(t)$ to zero, as long as intervals between change-points grow sufficiently fast with the horizon.

Non-stationarity detection in the piecewise stationary setting is a well-studied problem, with principled algorithmic solutions.  The non-stationary detection problem can be formulated as a quickest change detection (QCD) problem, for which methods have been developed that are optimal under various criteria  (see e.g. \citet{poor-hadj-qcd-book-2009}).

An important aspect of the piecewise stationary setting, is the nature of the change-point process. That is, the change-points can either be deterministic or random. In many works, an environment with deterministic change-points is considered \citep{liu2018change, cao2019nearly, auer2019adswitch, besson2022efficient}.
The analysis in this setting relies on the assumption that the change-points are not too close to each other, which is difficult to ensure for large horizons and leads to guarantees that are specific to this assumption \citep{besson2022efficient}. However, for practical scenarios, it is imperative to consider a more general model that allows for arbitrary change-point placement. Thus, we consider a random change-point model, which allows us to circumvent stringent assumptions (such as the one in \citep{besson2022efficient})  on the gaps between change-points in the analysis.
\begin{definition}
    The change-points are said to be \textit{i.i.d.} geometric with parameter $\eta>0$ if the intervals between consecutive change-points are \textit{i.i.d.} geometrically distributed with parameter $\eta$, i.e., $(\nu^{(k)} - \nu^{(k-1)}) \stackrel{\text{i.i.d.}}{\sim} \mathrm{Geo}(\eta)$. 
\end{definition}
Note that under geometric change-points, $\Delta(t)$ does not converge to $0$ as long as $\eta = \mathcal{O} \left(T^{-\xi} \right)$ for some $\xi \in (0,1)$. In the context of PS-MABs, the state-of-the-art work of \citet{besson2022efficient} has combined QCD methods with stationary bandit algorithms, achieving order optimal performance. It is important to note that this method relies on \emph{forced exploration}, i.e., the algorithm requires samples that are not obtained by the bandit procedure itself, in order to ensure that  changes are detected effectively.

\subsection{Random Restarting Algorithms}

Since MASTER's non-stationarity detection scheme fails for a critical range of values of the horizon $T$, MASTER effectively reduces to the MALG procedure from \citet{wei2021non} described in Section~\ref{subsec:MASTER}. MALG can be viewed as a random restarting algorithm with memory. The random restarts arise from random scheduling and switching when shorter instances exist, while the memory component is due to the possible reactivation without a  history reset. 

Seminal works on the NS-MAB problem and more  generally on non-stationary stochastic optimization problem have devised restarting procedures that use the knowledge of the degree of the non-stationarity to attain order optimal performance \citep{besbes_mab2014,besbes2015non}. Due to the inclusion of the geometric change-point model, we propose a simple random-restarting based procedure, termed as \emph{Random Restarts} (RR) that is order optimal, but \emph{not} prior-free, as a baseline for the setting of PS-MABs.

In the RR procedure, a bandit algorithm that is designed for the stationary setting is used as the base algorithm. A Bernoulli process with a suitably chosen success probability is used to decide when to reset and restart the base algorithm; therefore the times between restarts are i.i.d. geometric random variables.

\begin{theorem}\label{thm:rr_reg}
Let the geometric change-point parameter be $\eta = \Theta(T^{-\xi})$, for some $\xi\in(0,1)$. Set the intervals between random restarts of the RR algorithm to be i.i.d. $\mathrm{Geo}(\eta_{\mathrm{R}})$, for some $\eta_{\mathrm{R}}\in (0,1)$ with $\eta_{\mathrm{R}}>\eta$ (order-wise). If the regret of base algorithm is $\mathcal{O}(\mathrm{poly}(\log(T)))$, and $\eta_{\mathrm{R}}=\sqrt{\eta/\mathrm{poly}(\log(T))}$, the RR algorithm is order optimal for PS-MABs.
\end{theorem}
The proof of the theorem is included in the Appendix, along with a more detailed description of the RR algorithm. 
Even though RR is designed for geometric change-points, it can be appropriated to the setting of  deterministic change-points, by considering a fixed instance of the change-point process.


\section{EXPERIMENTAL STUDY}
\label{sec:experim}

\subsection{Setup And Algorithms}
The experiments were conducted on four different settings of piecewise stationarity, while keeping the same MAB structure. More specifically, we consider a PS-MAB problem with 5 arms and rewards that are bounded in $[0,1]$ in each step, for different values of the horizon $T$. In the simulations, we vary in the way the change-points are generated and in the effect of the changes that occur. Thus, we define the settings:

\textbf{Deterministic Change-Points.} For a given horizon $T$ and a number of change-points $N_C<T$, the change-points are (approximately) evenly spaced.

\textbf{Geometric Change-Points.} The change-points are geometric with parameter $\eta=T^{-\xi}, \xi\in(0,1)$.

\textbf{Uniform Problem.} To simulate a realistic scenario, in each change-point a uniformly selected number of arms have their means changed, and the magnitude of each change is uniformly selected in $[0.1,0.4]$.

\textbf{Worst-Case Problem.} For a worst-case scenario in terms of change detection, at each change, the arm with the lowest mean switches to the highest, while the other arms remain unchanged. Means are initialized at 0.3 plus a maximum difference of 0.005. At each change-point, the lowest mean changes to the highest mean plus a gap uniformly drawn from $[0.005, 0.05]$.

In every problem, if a mean exceeds $1$, the means are reset based on the initialization procedure of the problem. More details about the experimental setup are provided in the Appendix. 

\textbf{Algorithms.} We include the order optimal RR algorithm we propose, which is the only algorithm that has information about the non-stationarity. Since MASTER can be viewed as a random restarting procedure, as it does not have a functioning detection scheme, we employ an algorithm that randomly restarts the klUCB bandit algorithm \citep{garivier2011klucb} in each time step with probability $p=0.05$, termed as $\mathrm{RR}_{\mathrm{p.05}}$. Regarding the algorithms that employ QCD, we first include the state-of-the-art approach of \citet{besson2022efficient}, GLRklUCB, with the Bernoulli GLR change-detector. Finally, to demonstrate the robustness of the QCD-based methods, taking inspiration from GLRklUCB, we combine the Bernoulli GLR change-detector with klUCB and UCB \citep{auer2002finite}, due to its lower complexity compared to klUCB. However, unlike GLRklUCB, the latter algorithms do not use forced exploration, and we denote them as QCD$+$klUCB and QCD$+$UCB. More details on our QCD-based algorithms are given in the Appendix. 

\textbf{Knowledge Of $T$.} Although every method that is considered, except MASTER, implicitly assumes knowledge of the horizon $T$, they can easily be extended to unknown horizons through the use of the doubling trick \citep{besson2018doublingtrickscantmultiarmed}. 
%
MASTER essentially incorporates the doubling trick in the outer loop of the algorithm, by doubling the blocklength every time $n$ is updated. However, MASTER does not necessarily wait until the end of the block to update $n$ if Test 1 or Test 2 are triggered. Thus to ensure a fair and easy comparison with the other methods, we fix $n= \log_2 T$ throughout the operation of MASTER  and do not update $n$ and simply restart MALG if Test 1 or Test 2 are triggered.

\subsection{Experimental Results}
The experiments were conducted for horizons ranging from $T=1000$ to $T=100000$, with increments of approximately $5000$, averaged over $4000$ independent trials. Due to limited space, we present results for $T=100000$. For geometric change-points, we use $\xi \in \{0.3, 0.4, 0.5, 0.6, 0.7, 0.8\}$, and for deterministic change-points, $N_C \in \{3163, 1000, 317,$\\
$ 101, 32, 10\}$. The omitted results are available in the Appendix.

\subsubsection{Dynamic Regret}
\begin{figure*}[h]
\begin{center}
    \begin{subfigure}[b]{0.9\linewidth}
        \includegraphics[width=\linewidth]{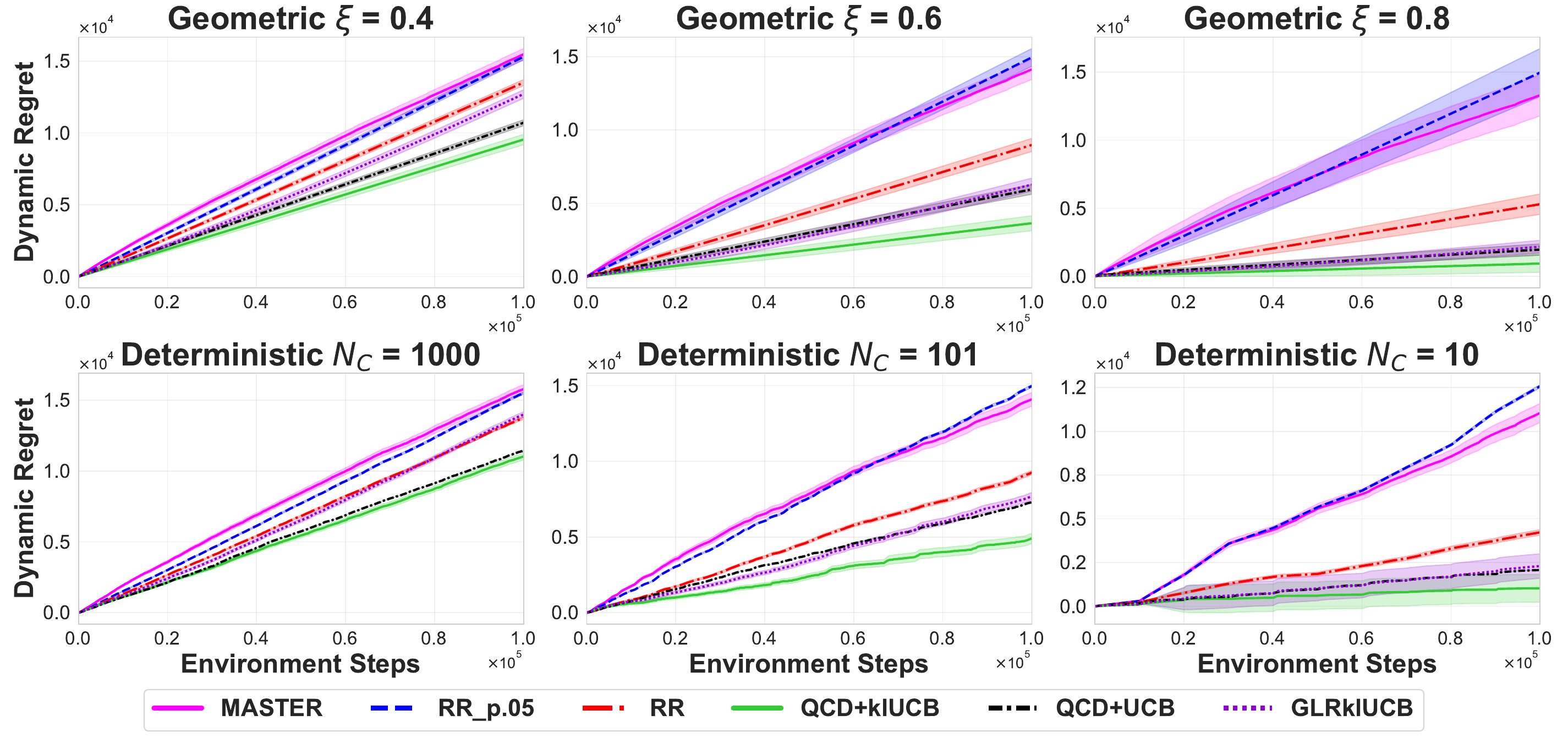}
        \caption{Uniform problem}
        \label{fig:reg_unif}
    \end{subfigure}
    \hfill
    \begin{subfigure}[b]{0.9\linewidth}
        \includegraphics[width=\linewidth]{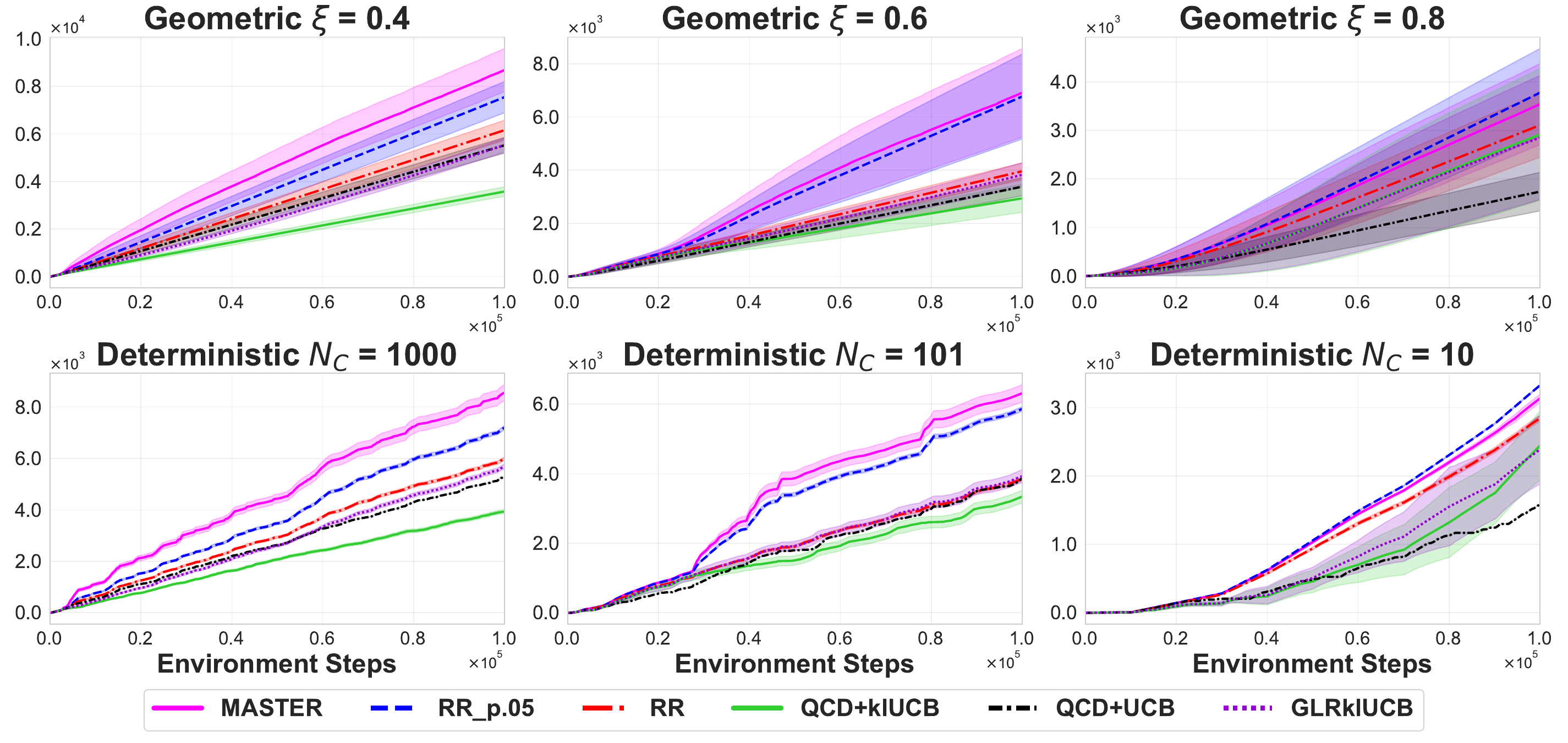}
        \caption{Worst-case problem}
        \label{fig:reg_worst}
    \end{subfigure}
\end{center}
\caption{Dynamic regret plots versus the time steps for $T=100000$, averaged over $4000$ independent runs. The case of the geometric change-points is on the top row for $\xi=0.4,0.6,0.8$ and the case of the deterministic change-points on the bottom row for $N_C=1000,101,10$. Left: Uniform problem. Right: Worst-case problem.}
\label{fig:reg_combined}
\end{figure*}

We compare the dynamic regret of the methods across time steps in the environment. As shown in Figure~\ref{fig:reg_unif}, for the Uniform problem, MASTER is consistently outperformed by RR. Additionally, the performance of $\mathrm{RR}_{\mathrm{p.05}}$ closely mirrors that of MASTER, reinforcing the observation about random restarting with memory. Regarding the QCD-based methods, GLRklUCB achieves order optimal performance, outperforming RR in nearly every instance except one, despite RR's knowledge of the non-stationarity. Moreover, GLRklUCB significantly surpasses both MASTER and $\mathrm{RR}_{\mathrm{p.05}}$. Surprisingly, $\mathrm{QCD}+$ not only outperforms MASTER but also exceeds the performance of RR and GLRklUCB. A similar trend is evident in the Worst-Case problem (Figure \ref{fig:reg_worst}), where MASTER is outperformed by every method, including $\mathrm{RR}_{\mathrm{p.05}}$ at higher levels of non-stationarity. Across both problems, the QCD-based methods consistently demonstrate the best performance.

\begin{figure}[h]
\begin{center}
\includegraphics[width=0.82\linewidth]{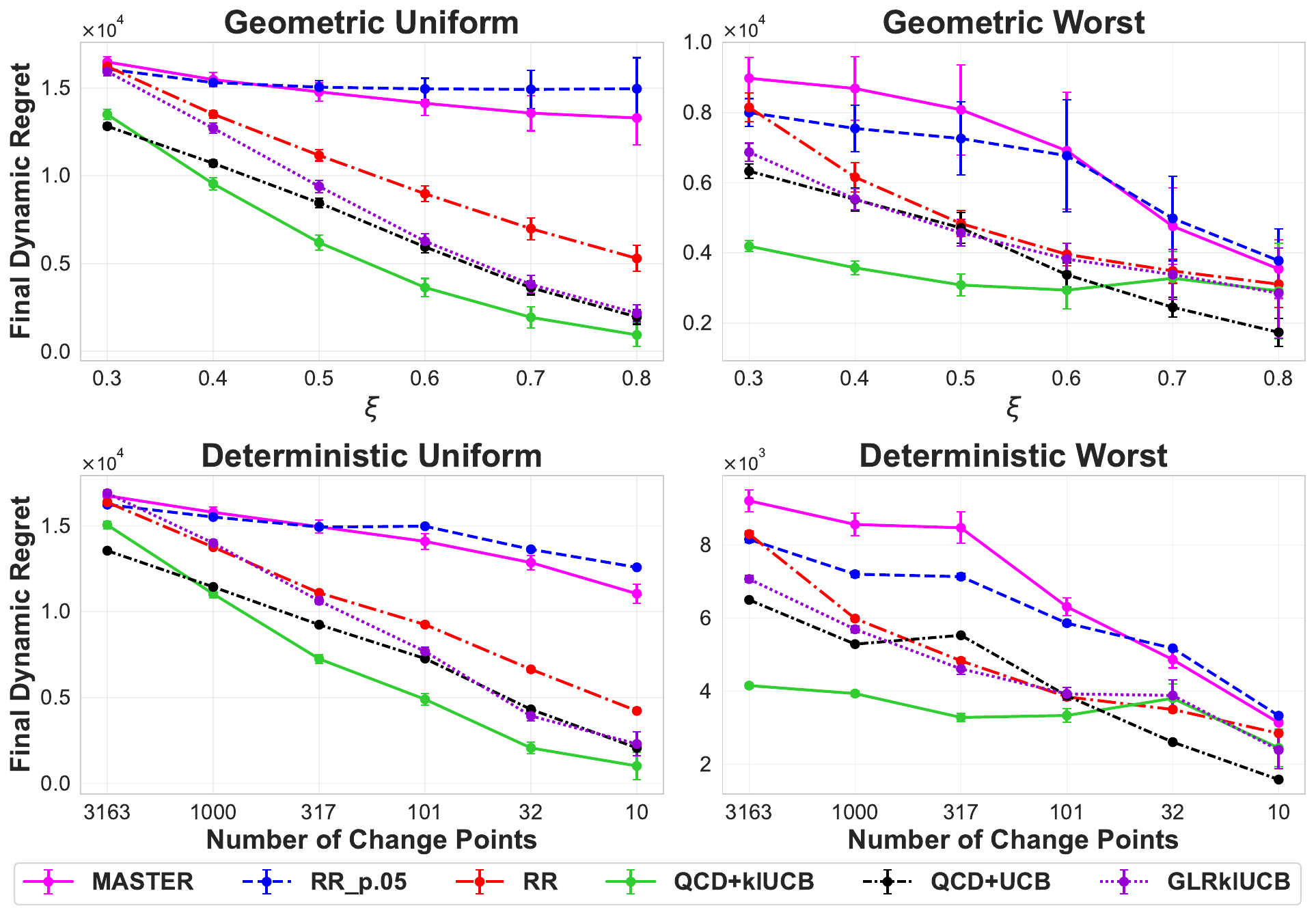}
\end{center}
\caption{Final dynamic regret versus the decrease in non-stationarity, for $T=100000$, $4000$ runs.}
\label{fig:robust_plot}
\end{figure}

To assess the robustness of the methods, we plot the final dynamic regret ($\mathrm{Reg}_{\mathrm{D}}(T)$) as the non-stationarity decreases. A robust method should exhibit lower final dynamic regret with decreasing non-stationarity, while maintaining small variance. As shown in Figure \ref{fig:robust_plot}, MASTER's robustness is comparable to that of $\mathrm{RR}_{\mathrm{p.05}}$, and it is significantly worse than that of RR. In contrast, the QCD-based methods demonstrate superior robustness compared to MASTER and the random restarting algorithms. The final regret of the QCD-based methods display lower variance than that of MASTER and $\mathrm{RR}_{\mathrm{p.05}}$, and in certain cases, lower than RR. Thus, the QCD-based methods show greater robustness, achieving better regret performance and lower variance as the non-stationarity decreases. Among them, QCD+UCB achieves the lowest variance.

\subsubsection{Detection And Time Efficiency}
To verify our theory, we examine the non-stationarity detection mechanism of MASTER. As seen in Table \ref{table:change_detect}, MASTER indeed does not declare any changes for any of the problems, confirming the result in Theorem~\ref{thrm:theor_1}. Conversely, the QCD-based approaches are able to declare a significant number of the change-points, and enforce restarting with small delay after the change-point. 


Lastly, we evaluate the computational efficiency of the algorithms based on the time required per run. As shown in Table  \ref{table:time_per_exp}, MASTER is less efficient compared to the QCD-based methods. Due to its similar performance with MASTER, we include $\mathrm{RR}_{\mathrm{p.05}}$, which is 2 to 3 times more efficient than the former. Notably, using the klUCB algorithm increases the complexity of both the random-restarting and the QCD-based algorithms, resulting in longer run times. In contrast, opting for the UCB algorithm significantly reduces the runtime, offering speed improvements over MASTER by a factor of 5 to 37 as non-stationarity increases. 

\begin{table}[H]
\caption{Mean Number Of Declared Changes For Different Change-Points, $T=100000$, 4000 Runs.}
\label{table:change_detect}
\scriptsize	 
\setlength{\tabcolsep}{3pt} 
\begin{center}
\begin{tabular}{c|c|cccccc}
\multicolumn{8}{c}{\textbf{(a) Random Change-Points (Columns: $\xi$)}} \\ \hline
\multicolumn{1}{c|}{\bf Algorithm} & \multicolumn{1}{c|}{\bf PB} & \multicolumn{1}{c}{\bf 0.3} & \multicolumn{1}{c}{\bf 0.4} & \multicolumn{1}{c}{\bf 0.5} & \multicolumn{1}{c}{\bf 0.6} & \multicolumn{1}{c}{\bf 0.7} & \multicolumn{1}{c}{\bf 0.8} \\
\hline
{MASTER} & {Unif} & 0.00 & 0.00 & 0.00 & 0.00 & 0.00 & 0.00 \\
{MASTER} & {Worst} & 0.00 & 0.00 & 0.00 & 0.00 & 0.00 & 0.00 \\
{GLRklUCB} & {Unif} & 129.43 & 137.21 & 99.21 & 54.05 & 23.78 & 8.88 \\
{GLRklUCB} & {Worst} & 147.03 & 79.08 & 45.28 & 20.73 & 6.07 & 1.46 \\
{QCD+UCB} & {Unif} & 121.62 & 139.51 & 104.03 & 58.01 & 25.35 & 9.23 \\
{QCD+UCB} & {Worst} & 150.80 & 110.35 & 79.12 & 36.51 & 11.77 & 3.77 \\
{QCD+klUCB} & {Unif} & 134.84 & 131.53 & 91.32 & 48.76 & 21.53 & 8.22 \\
{QCD+klUCB} & {Worst} & 131.94 & 64.20 & 31.33 & 12.93 & 4.59 & 1.27 \\
\hline
\end{tabular}

\vspace{0.5em} 

\begin{tabular}{c|c|cccccc}
\multicolumn{8}{c}{\textbf{(b) Deterministic Change-Points (Columns: $N_C$)}} \\ \hline
\multicolumn{1}{c|}{\bf Algorithm} & \multicolumn{1}{c|}{\bf PB} & \multicolumn{1}{c}{\bf 3163} & \multicolumn{1}{c}{\bf 1000} & \multicolumn{1}{c}{\bf 317} & \multicolumn{1}{c}{\bf 101} & \multicolumn{1}{c}{\bf 32} & \multicolumn{1}{c}{\bf 10} \\
\hline
{MASTER} & {Unif} & 0.00 & 0.00 & 0.00 & 0.00 & 0.00 & 0.00 \\
{MASTER} & {Worst} & 0.00 & 0.00 & 0.00 & 0.00 & 0.00 & 0.00 \\
{GLRklUCB} & {Unif} & 98.45 & 136.94 & 119.82 & 69.49 & 27.38 & 8.31 \\
{GLRklUCB} & {Worst} & 148.35 & 83.78 & 51.12 & 21.52 & 6.74 & 1.19 \\
{QCD+UCB} & {Unif} & 84.00 & 149.00 & 122.00 & 83.00 & 28.00 & 9.00 \\
{QCD+UCB} & {Worst} & 141.00 & 95.00 & 102.00 & 36.00 & 13.00 & 4.00 \\
{QCD+klUCB} & {Unif} & 104.45 & 134.02 & 113.89 & 62.68 & 25.81 & 7.95 \\
{QCD+klUCB} & {Worst} & 130.42 & 71.08 & 34.60 & 14.37 & 4.38 & 1.11 \\
\hline
\end{tabular}
\end{center}
\end{table}

\begin{table*}[h]
\caption{Average Time Per Run (sec) For Different Change-Points, $T$=100000, 4000 Runs.}
\label{table:time_per_exp}
\scriptsize
\setlength{\tabcolsep}{3pt} 
\begin{center}
\begin{tabular}{c|c|cccccc}
\multicolumn{8}{c}{\textbf{(a) Random Change-Points}} \\ \hline
\multicolumn{1}{c|}{\bf Algorithm} & \multicolumn{1}{c|}{\bf PB} & \multicolumn{1}{c}{\bf $\xi$=0.3} & \multicolumn{1}{c}{\bf $\xi$=0.4} & \multicolumn{1}{c}{\bf $\xi$=0.5} & \multicolumn{1}{c}{\bf $\xi$=0.6} & \multicolumn{1}{c}{\bf $\xi$=0.7} & \multicolumn{1}{c}{\bf $\xi$=0.8} \\
\hline
{MASTER} & {Unif} & {3.33 ± 0.03} & {3.34 ± 0.02} & {3.34 ± 0.04} & {3.33 ± 0.03} & {3.36 ± 0.04} & {3.35 ± 0.20} \\
{MASTER} & {Worst} & {3.32 ± 0.03} & {3.32 ± 0.01} & {3.33 ± 0.02} & {3.33 ± 0.05} & {3.35 ± 0.08} & {3.36 ± 0.15} \\
{RR\_{p.05}} & {Unif} & {1.13 ± 0.02} & {1.12 ± 0.01} & {1.12 ± 0.02} & {1.12 ± 0.03} & {1.12 ± 0.05} & {1.12 ± 0.07} \\
{RR\_{p.05}} & {Worst} & {0.96 ± 0.01} & {0.94 ± 0.02} & {0.94 ± 0.03} & {0.96 ± 0.05} & {1.03 ± 0.09} & {1.23 ± 0.05} \\
{GLRklUCB} & {Unif} & {1.38 ± 0.01} & {1.37 ± 0.01} & {1.45 ± 0.02} & {1.65 ± 0.05} & {2.18 ± 0.20} & {3.76 ± 1.02} \\
{GLRklUCB} & {Worst} & {1.33 ± 0.03} & {1.49 ± 0.04} & {1.68 ± 0.11} & {2.14 ± 0.37} & {3.27 ± 0.53} & {6.08 ± 2.06} \\
{QCD+UCB} & {Unif} & {0.09 ± 0.01} & {0.08 ± 0.01} & {0.11 ± 0.01} & {0.24 ± 0.04} & {0.72 ± 0.20} & {2.24 ± 1.04} \\
{QCD+UCB} & {Worst} & {0.08 ± 0.01} & {0.10 ± 0.01} & {0.16 ± 0.02} & {0.33 ± 0.07} & {0.68 ± 0.19} & {1.65 ± 0.71} \\
{QCD+klUCB} & {Unif} & {1.68 ± 0.01} & {1.68 ± 0.01} & {1.73 ± 0.01} & {1.90 ± 0.04} & {2.44 ± 0.22} & {4.12 ± 1.09} \\
{QCD+klUCB} & {Worst} & {1.67 ± 0.01} & {1.80 ± 0.06} & {2.06 ± 0.28} & {2.71 ± 0.84} & {3.60 ± 0.55} & {6.42 ± 2.13} \\
\hline
\end{tabular}

\vspace{0.5em}

\begin{tabular}{c|c|cccccc}
\multicolumn{8}{c}{\textbf{(b) Deterministic Change-Points}} \\ \hline
\multicolumn{1}{c|}{\bf Algorithm} & \multicolumn{1}{c|}{\bf PB} & \multicolumn{1}{c}{\bf $N_C$=3163} & \multicolumn{1}{c}{\bf $N_C$=1000} & \multicolumn{1}{c}{\bf $N_C$=317} & \multicolumn{1}{c}{\bf $N_C$=101} & \multicolumn{1}{c}{\bf $N_C$=32} & \multicolumn{1}{c}{\bf $N_C$=10} \\
\hline
{MASTER} & {Unif} & {3.35 ± 0.42} & {3.32 ± 0.04} & {3.35 ± 0.22} & {3.32 ± 0.01} & {3.32 ± 0.15} & {3.32 ± 0.02} \\
{MASTER} & {Worst} & {3.38 ± 0.09} & {3.38 ± 0.10} & {3.40 ± 0.08} & {3.38 ± 0.06} & {3.37 ± 0.03} & {3.35 ± 0.07} \\
{RR\_{p.05}} & {Unif} & {1.15 ± 0.26} & {1.14 ± 0.04} & {1.16 ± 0.12} & {1.12 ± 0.01} & {1.10 ± 0.04} & {1.08 ± 0.11} \\
{RR\_{p.05}} & {Worst} & {0.97 ± 0.05} & {0.95 ± 0.05} & {0.94 ± 0.02} & {1.00 ± 0.01} & {1.04 ± 0.01} & {1.24 ± 0.01} \\
{GLRklUCB} & {Unif} & {1.46 ± 0.28} & {1.38 ± 0.03} & {1.43 ± 0.17} & {1.52 ± 0.01} & {1.83 ± 0.06} & {2.67 ± 0.20} \\
{GLRklUCB} & {Worst} & {1.33 ± 0.06} & {1.48 ± 0.07} & {1.66 ± 0.06} & {2.04 ± 0.11} & {2.97 ± 0.17} & {5.06 ± 0.94} \\
{QCD+UCB} & {Unif} & {0.12 ± 0.08} & {0.07 ± 0.01} & {0.13 ± 0.05} & {0.13 ± 0.00} & {0.38 ± 0.04} & {1.09 ± 0.14} \\
{QCD+UCB} & {Worst} & {0.09 ± 0.02} & {0.11 ± 0.03} & {0.16 ± 0.02} & {0.41 ± 0.01} & {0.73 ± 0.00} & {1.05 ± 0.03} \\
{QCD+klUCB} & {Unif} & {1.71 ± 0.29} & {1.69 ± 0.03} & {1.73 ± 0.17} & {1.78 ± 0.01} & {2.02 ± 0.07} & {2.88 ± 0.17} \\
{QCD+klUCB} & {Worst} & {1.68 ± 0.06} & {1.77 ± 0.07} & {1.95 ± 0.07} & {2.35 ± 0.16} & {3.24 ± 0.22} & {5.09 ± 0.77} \\
\hline
\end{tabular}
\end{center}
\end{table*}

\section{SUMMARY AND OUTLOOK}
In this work, we examined the feasibility of prior-free black-box NS-RL in practice, focusing on the only existing method that is order optimal: MASTER. We showed that there is a large span of horizon values (up to at least 1.24 billion) where MASTER's non-stationarity detection mechanism is ineffective, reducing its performance to that of a random restarting approach (with some memory). Furthermore, we demonstrated that the performance bounds for MASTER given in \citet{wei2021non} are not useful for an unreasonably large range of horizon values (up to a horizon of at least $4 \times 10^{14}$). To validate these findings, we considered the special NS-RL setting of PS-MABs, incorporating both random and deterministic change-points. To establish a suitable random restarting baseline for our experiments, we proposed an order optimal random restarting algorithm tailored for PS-MABs. Our experiments confirmed MASTER's limitations, while showing that prior-free methods utilizing QCD-based mechanisms effectively detect changes, yielding superior performance and robustness. 

These observations highlight a gap in prior-free, black-box NS-RL: the need for an order optimal algorithm that is not only theoretically sound but also practically feasible. 
The theoretical analysis of MASTER provided in \citet{wei2021non} is a valuable starting point, and future work could focus on developing approaches that mimic MASTER's multi-scale guarantees. On the other hand, the insight that effective non-stationarity detection is key to good performance, paves the way for developing principled change detection algorithms, based on QCD theory, that can be generalized to broader NS-RL settings.

\section{ACKNOWLEDGEMENT}
We would like to thank Subhonmesh Bose for the insightful discussions.

\vfill
\pagebreak

\appendix
\section{ALGORITHMIC IMPLEMENTATIONS}
\subsection{The MASTER Algorithm}

\begin{algorithm}
\caption{\textbf{M}ulti-scale \textbf{A}lgorithm with \textbf{S}tationary \textbf{TE}sts and \textbf{R}estarts}\label{alg:CDB}
\textbf{Input}: $\rho (\cdot)$, input algorithm \textsf{ALG}. \\
\textbf{Initialization}: $t \leftarrow 1$.

\begin{algorithmic}[1]
\FOR{$n=0,1,\dots$}
    \STATE $t_{n} \leftarrow t$.
    \FOR{$m = 0, 1, \dots, n$}
        \FOR{$k = 0, 1, \dots, 2^{n-m}-1$}
            \STATE $I_{k} \leftarrow [t_{n} + k2^{m}, \dots, t_{n} + (k+1)2^{m}-1]$
            \STATE With probability $\frac{\rho(2^{n})}{\rho(2^{m})}$, schedule an instance of \textsf{ALG}, $alg, $ to $I_{k}$\\
            with start $alg.s = t_{n} + k2^{m}$ and end $alg.e = t_{n} + (k+1)2^{m}-1$.
        \ENDFOR
    \ENDFOR
    \FOR{$t = t_{n}, t_{n} + 1, \dots, t_{n} + 2^{n} - 1$}
        \IF{$t > T$}
            \STATE End the algorithm.
        \ENDIF
        \STATE Let $alg$ be the instance assigned to the shortest $I_k$, for all $k$, that contains $t$.
        \STATE Receive $\tilde{f}_{t}$ and $\pi_{t}$ from $alg$.
        \STATE Execute $\pi_{t}$ and receive the reward $R_{t}$.
        \STATE $\tilde{g}_{t} \leftarrow \tilde{f}_{t}$.
        \IF{(Test 1) $t = alg'.e$ for some $2^m$-length instance $alg'$ and $\frac{1}{2^m}\sum_{\tau = alg'.s}^{alg'.e} R_\tau -\min_{\tau\in[t_n,t]} \tilde{g}_{\tau}\geq 54(\log_2T +1)\log(T/\delta)\rho(2^m)$}
            \STATE Non-stationarity detected. Restart from Line 1.
        \ENDIF
        \IF{(Test 2) $\frac{1}{t-t_n+1}\sum_{\tau=t_n}^t (\tilde{g}_\tau -R_\tau)\geq 18(\log_2T +1)\log(T/\delta)\rho(t-t_n+1)$}
            \STATE Non-stationarity detected. Restart from Line 1.
        \ENDIF
    \ENDFOR
\ENDFOR
\end{algorithmic}
\end{algorithm}

From Line 2 to Line 8, MASTER first partitions the  interval of length $2^{n}$ into $2^{n-m}$ sub-intervals of length $2^{m}$, which are denoted by $I_{k}$. Then, each sub-interval is assigned an instance of \textsf{ALG} with probability $\frac{\rho(2^{n})}{\rho(2^{m})}$ independently. Note that there is always an instance of \textsf{ALG} assigned to the original interval of length $2^{n}$ since $\frac{\rho(2^{n})}{\rho(2^{n})} = 1$. At Line 13, MASTER selects the instance with the shortest interval to run and pauses all the other instances. Therefore, whenever there is a shorter instance compared to the current active one at the next time-step, MASTER makes the current instance inactive and the shorter one active. If any previously active instance becomes the one with the smallest length, it resumes and runs with the data prior to the pause. Thus, the MASTER algorithm is a restarting scheme with memory. Moreover, MASTER restarts if it reaches the end of the $2^n$ block, and then $n$ is incremented by $1$ for the next block. MASTER can also restart through Test 1 and 2, which are used to detect non-stationarity. 
For a more detailed discussion regarding the MASTER algorithm, we refer the reader to \citet{wei2021non}.

\subsection{Multi-Armed Bandit Preliminaries}
In order to present the different algorithms for Piecewise Stationary Multi-Armed Bandits (PS-MABs), we first provide some notation that will be used in the context of MABs.

We assume that a (stationary) bandit algorithm, $\mathcal{B}$, maintains a history list for its arms, which includes the number of pulls for each arm, along with the rewards that are obtained from each pull. We denote this history by $\mathcal{H}_{\mathcal{B}}$. Based on Definition \ref{def:ns_env}, we have that the policy set $\Pi$ corresponds to the set of arms, which is given by $\{1,\dots,A\}$ when we have a total of $A$ arms. Finally, we use the notation $\mathcal{H}_{\mathcal{B},a}$ for any arm $a\in\{1,\dots,A\}$ to denote the bandit history that corresponds to arm $a$.

\subsection{The RR Algorithm}
In this section, we present the Random Restarts (RR) algorithm for PS-MABs, which is given as follows.

\begin{algorithm}
\caption{Random Restarts}\label{alg:RR}
\textbf{Input}: base bandit algorithm $\mathcal{B}$ with $\mathrm{poly}(\log (T))$ regret, number of arms $A$, geometric distribution parameter $\eta$, horizon $T$ \\
\textbf{Initialization}: set random restart intervals as i.i.d. $\mathrm{Geo}(\sqrt{\eta/\mathrm{poly}(\log (T))})$, history list $\mathcal{H}_{\mathcal{B}}\leftarrow\emptyset$.

\begin{algorithmic}[1]
\FOR{$t=1,2,\dots,T$}
    \IF{$t$ is a random restart point}
        \STATE $\mathcal{H}_{\mathcal{B}}\leftarrow\emptyset$.
    \ENDIF
    \STATE $A_t\leftarrow \mathcal{B}(\mathcal{H}_{\mathcal{B}})$ (Bandit Algorithm).
    \STATE Pull arm $A_t$ and receive reward $R_t$.
    \STATE Add $(A_t,R_t)$ into $\mathcal{H}_{\mathcal{B}}$.
\ENDFOR
\end{algorithmic}
\end{algorithm}

\subsection{The QCD+ Algorithm}
Before we present the algorithm, we introduce the notion of a change detector $\mathcal{D}$. The change detector takes as input the bandit history for a specific arm $a$, $\mathcal{H}_{\mathcal{B},a}$, and yields ``True" if, according to its test, there has been a change in that specific arm so far, and yields ``False" otherwise. The QCD+ algorithm is an adaptation of the GLRklUCB algorithm, proposed in \citet{besson2022efficient}, and is obtained by removing the forced exploration and the local restart conditions. Thus, the algorithm is given as follows.
\begin{algorithm}[H]
\caption{QCD+}\label{alg:QCD+}
\textbf{Input}: change detector $\mathcal{D}$, bandit algorithm $\mathcal{B}$,  horizon $T$, number of arms $A$ \\
\textbf{Initialization}: history list $\mathcal{H}_{\mathcal{B}}\leftarrow\emptyset$.

\begin{algorithmic}[1]
\FOR{$t=1,2,\dots,T$}
        \STATE $A_t\leftarrow \mathcal{B}(\mathcal{H}_{\mathcal{B}})$ (Bandit Algorithm).
    \STATE Pull arm $A_t$ and receive reward $R_t$.
    \STATE Add $(A_t,R_t)$ into $\mathcal{H}_{\mathcal{B}}$.
    \IF{$\mathcal{D}\left( \mathcal{H}_{\mathcal{B},A_t}\right)=\mathrm{True}$ (Change Detected)}
        \STATE $\mathcal{H}_{\mathcal{B}}\leftarrow\emptyset$.
    \ENDIF
\ENDFOR
\end{algorithmic}
\end{algorithm}
\section{THEORETICAL PROOFS}
\setcounter{theorem}{0}
\setcounter{lemma}{0}
\setcounter{definition}{0}
\setcounter{assumption}{0}
\subsection{Theory Preliminaries}
To enhance the clarity and flow of the proofs, we include key points from the main paper.

\begin{definition} Assume an agent that interacts with an environment through a behavior-policy set $\Pi$ over a horizon $T$, that is, a fixed number of time steps. The environment has a collection of $T$ reward functions $f_1,...,f_T:\Pi \rightarrow [0,1]$, unknown to the agent, that model the environment's feedback. In each time step $t=1,...,T$, the agent selects a policy $\pi_t\in \Pi$ and receives the feedback in the form of a random reward $R_t\in [0,1]$ with mean $f_t(\pi_t)=\mathbb{E}[R_t|\pi_t]$. The goal is to minimize the dynamic regret,
    $\mathrm{Reg}_{\mathrm{D}}(T)=\mathbb{E}\left[\sum_{t=1}^T (f_t^*-f_t(\pi_t))\right]$, where $f_t^*=\max_{\pi\in\Pi} f_t(\pi)$ is the best expected reward at time $t$.
\end{definition}

\begin{definition}
   (Following \citet{wei2021non}.) $\Delta:[T]\rightarrow \mathbb{R}$ is a non-stationarity measure, if it satisfies $\Delta(t)\geq \max_{\pi \in \Pi} |f_t(\pi)-f_{t+1}(\pi)|$,  for all $t$. For interval $[s,e]$, the cumulative non-stationarity is given as $\Delta_{[s,e]}=\sum_{\tau=s}^{e-1} \Delta(\tau)$, with $\Delta_{[s,s]}=0$, and the number of changes ($+$1) as $L_{[s,e]}=1+\sum_{\tau=s}^{e-1}\mathds{1}{[\Delta(\tau)\neq 0]}$.
\end{definition}

\begin{assumption}
(Following \citet{wei2021non}.) ALG outputs $\tilde{f}_t\in [0,1]$ at the beginning of each time step $t$. Suppose there exists a non-increasing function $\rho: [T] \rightarrow \mathbb{R}$ such that $\rho(t) \geq 1/\sqrt{t}$ and $C(t) = t \rho(t)$ is non-decreasing. Additionally, assume $\Delta(t)$ to be a non-stationary measure. Then, for any $t \in [T]$ such that $\Delta_{[1,t]} \leq \rho(t)$, the following holds with probability at least $1-\delta/T$:
\begin{enumerate}
    \item $\tilde{f}_t \geq \min _{\tau\in[1,t]}f^*_\tau - \Delta_{[1,t]}$, and
    \item $\frac{1}{t}\sum_{\tau=1}^t\left( \tilde{f}_\tau - R_\tau \right)\leq \rho(t) + \Delta_{[1,t]}$.
\end{enumerate}
\end{assumption}

\begin{lemma}(Adapted from \citet{wei2021non}.)
Let \(\hat{n} = \log_2 T + 1\), \(\hat{\rho}(t) = 6\hat{n} \log(T/\delta)\rho(t)\) and \(t' = t - alg.s + 1\). Suppose Assumption \ref{assumption_one} holds for ALG, and that \(n \leq \log_2 T\). Then, for any instance $alg$, with start at $alg.s$ and finish at $alg.e$, that is maintained by MALG and any \(t \in [alg.s, alg.e]\) that satisfies $\Delta_{[alg.s,t]} \leq \rho(t')$, MALG satisfies with with probability at least $1-\delta/T$:
\begin{enumerate}
    \item $\tilde{g}_t \geq \min_{\tau \in [alg.s,t]} f^*_\tau - \Delta_{[alg.s,t]}$, and
    \item $\frac{1}{t'} \sum_{\tau = alg.s}^t (\tilde{g}_\tau - R_\tau) \leq \hat{\rho}(t') + \hat{n} \Delta_{[alg.s,t]}$.
\end{enumerate}
\end{lemma}

Finally, we restate the soft-Oh notation, $\tilde{\mathcal{O}}(\cdot)$, and introduce the soft-Omega notation, $\tilde{\Omega}(\cdot)$. We write $h_1(x)=\Tilde{\mathcal{O}}(h_2(x))$ with high probability, if $h_1(x)=\mathcal{O}(\mathrm{poly}(\log(T/\delta))h_2(x))$ with probability $1-\delta$ for some $\delta\in (0,1)$. On the other hand, $h_1(x)=\Tilde{\Omega}(h_2(x))$ with high probability, if $h_1(x)=\Omega(\mathrm{poly}(\log(T/\delta))h_2(x))$ with probability $1-\delta$ for some $\delta\in (0,1)$

\subsection{Proof Of Lemma \ref{lemma:test_work}}
\setcounter{lemma}{1}
\begin{lemma}
    Test 1 will not be triggered for any value of the horizon that satisfies $\rho(T)>\frac{1}{54(\log_2T +1)\log(T/\delta)}$ and Test 2 will also not be triggered if $\rho(T)>\frac{1}{18(\log_2T +1)\log(T/\delta)}$.
\end{lemma}
\textbf{Proof}:
    
     According to Test 1, in order to detect non-stationarity, the following condition needs to hold,
         \begin{equation}
             \frac{1}{2^m}\sum_{\tau=alg.s}^{alg.e}R_\tau -\min_{\tau\in[t_n,t]} \tilde{g}_{\tau}\geq 54(\log_2T +1)\log(T/\delta)\rho(2^m).\label{eq:test_1_appendix}
         \end{equation}
         To effectively detect using Test 1, we need to ensure that the LHS in Equation \eqref{eq:test_1_appendix} is indeed greater than the RHS. Thereupon, when the LHS is maximized and the RHS is minimized, Test 1 must always work. By definition, $\tilde{g}_\tau$ is the active instance of $\tilde{f}_\tau$   at round $\tau$, and hence $\tilde{g}_\tau\in [0,1]$, while we have that $R_\tau\in [0,1]$. Additionally, since $\rho(t)$ is a non-increasing function and because $m=0,1,\dots,n$, its minimum value is achieved when $m=n$. Therefore, in the most ideal detection scenario, which is not necessarily met in practice, the following needs to hold,
         \begin{equation*}
             \frac{1}{2^m}\sum_{\tau=alg.s}^{alg.e}1 \geq 54(\log_2T +1)\log(T/\delta)\rho(2^n).
         \end{equation*}
         Note that if this condition is not met, then, with absolute certainty, no non-stationarity will be detected by Test 1. Finally, since the largest value that $n$ can have, to ensure the performance guarantees of MASTER, is equal to $\log_2T$, we can minimize the RHS even further. In particular, since the $alg$ instance has length equal to $2^m$,
         \begin{equation*}
             1 \geq 54(\log_2T +1)\log(T/\delta)\rho(2^{\log_2 T}),
         \end{equation*}
         which implies,
         \begin{equation}
             \rho(T)\leq \frac{1}{54(\log_2 T +1)\log(T/\delta)}.
             \label{eq:Test1_detect_cond}
         \end{equation}
         Regarding Test 2, the same idea as before is applied, 
         \begin{equation*}
              \frac{1}{t-t_n+1}\sum_{\tau=t_n}^t (\tilde{g}_\tau -R_\tau)\geq 18(\log_2T +1)\log(T/\delta)\rho(t')
         \end{equation*}
         Since $\tilde{g}_{\tau}, R_{\tau} \in [0,1]$ for any $\tau \in \{1,\dots,T\}$, we have
         \begin{equation*}
              \frac{1}{t-t_n+1}\sum_{\tau=t_n}^t  1 = 1 \geq 18(\log_2T +1)\log(T/\delta)\rho(2^n).
         \end{equation*}
         Thus, we have
         \begin{equation}
              \rho(T)\leq \frac{1}{18(\log_2T +1)\log(T/\delta)} .
              \label{eq:Test2_detect_cond}
         \end{equation}
Hence, if the opposite of the inequalities \eqref{eq:Test1_detect_cond} and \eqref{eq:Test2_detect_cond} hold, neither Test 1 nor Test 2 will be triggered for any value of the horizon, which concludes the proof.
\subsection{Proof Of Theorem \ref{thrm:theor_1}}
We first restate the theorem.
\begin{theorem}
    If Assumption \ref{assumption_one} is satisfied, the thresholds of Test 1 and Test 2 can be crossed only if $\delta\geq T \exp\left(- \frac{\sqrt{T}}{54(\log_2T+1)} \right)$, and since $\delta<1$, $T$ must be at least 1.24 billion.
\end{theorem}
\textbf{Proof}:
    
        According to Lemma \ref{lemma:test_work}, in order to ensure that Test 1 and Test 2 can be triggered, the following conditions need to hold:
        \begin{equation*}
        \rho(T)\leq \frac{1}{54(\log_2T +1)\log(T/\delta)}\quad \mbox{and}\quad \rho(T)\leq \frac{1}{18(\log_2T +1)\log(T/\delta)}.
        \end{equation*}
        Based on Assumption \ref{assumption_one}, the function $\rho$ needs to satisfy, $\rho(t) \geq 1/\sqrt{t}$. Thereupon, it is evident that if the following conditions hold, then Test 1 and Test 2 cannot be triggered:
        \begin{equation*}
             \frac{1}{\sqrt{T}}> \frac{1}{54(\log_2T +1)\log(T/\delta)}\quad \mbox{and}\quad \frac{1}{\sqrt{T}}> \frac{1}{18(\log_2T +1)\log(T/\delta)}
        \end{equation*}
        Solving with respect to $\delta$, we have that,
        \begin{equation*}
             \delta< T \exp\left(- \frac{\sqrt{T}}{54\log_2(T+1)} \right)\quad \mbox{and}\quad \delta< T \exp\left(- \frac{\sqrt{T}}{18\log_2(T+1)} \right).
        \end{equation*}
        Thereupon, in order to ensure that both tests can be triggered, the following must hold:
         \begin{equation*}
             \delta \geq T \exp\left(- \frac{\sqrt{T}}{54\log_2(T+1)} \right)\quad \mbox{and}\quad \delta \geq T \exp\left(- \frac{\sqrt{T}}{18\log_2(T+1)} \right).
        \end{equation*}
Thus, by selecting the more constrained inequality, the first part of the theorem follows.

Thereupon, the following inequality needs to hold,
\begin{equation*}
    \delta \geq T \exp\left(- \frac{\sqrt{T}}{54\log_2(T+1)} \right).
\end{equation*}
However, recall that the for the probabilistic parameter $\delta\in(0,1)$. Hence, since $\delta$ is trivially upper-bounded by 1, the following needs to hold,
\begin{equation*}
    1 > T \exp\left(- \frac{\sqrt{T}}{54\log_2(T+1)} \right)
\end{equation*}
which we can rearrange as
\begin{equation}
    \exp\left( \frac{\sqrt{T}}{54\log_2(T+1)} \right) > T .
    \label{eq:Test_12_work_req}
\end{equation}
To make sure that either Test 1 or Test 2 can indeed detect the non-stationarity, we need to guarantee that the LHS in \eqref{eq:Test_12_work_req} is greater than the RHS in \eqref{eq:Test_12_work_req}. Solving this inequality numerically, we get that $T$ needs to be at least equal to $1246257461$, which gives the second part of the theorem.

\setcounter{corollary}{1}
\subsection{Proof Of Corollary \ref{cor:master_perf}}
We first restate Corollary.
\begin{corollary}
    If Assumption \ref{assumption_one} holds, then the $\mathrm{Reg}_{\mathrm{D}}(T)$ bound of MASTER is at least \[
    B_\mathrm{D}(T)=24 (\log_2(T) + 1) \log (T) \sqrt{T} \left( 1 + 15 \log(T) \right).\] Therefore, whenever
    \[T\leq 4\times 10^{14},\quad T < B_\mathrm{D}(T),\] i.e., the $\mathrm{Reg}_{\mathrm{D}}(T)$ bound of MASTER is trivial.
\end{corollary}
\textbf{Proof}:

To prove the corollary, we follow the approach used in the proof of the regret bound for MASTER, as presented in Theorem 2 of \citet{wei2021non}. The proof begins by deriving a regret bound for any length-$2^n$ block, which is then tailored to a specific form of $C(t)$. Since MASTER employs the two non-stationarity detection tests that can trigger restarts, the regret between two restarts is appropriately upper bounded using the refined form of the length-$2^n$ block regret bound. Thus, the regret bound for any length-$2^n$ block has to be tight, otherwise, the final regret bound would be sub-optimal. This bound essentially serves as the foundational bound for the algorithm, with the final regret bound constructed from the exact or upper bounds of its components.

The regret bound for any length-$2^n$ block is provided in Lemma 4 from \citet{wei2021non} and its proof is given in Lemma 16 from \citet{wei2021non}. We present Lemma 16 from \citet{wei2021non} below.

\begin{lemma}
(Adapted from \citet{wei2021non}.) Let the high-probability events described in Lemma \ref{lemma_3_master} hold. Then with high probability, the following holds for any length-$2^n$ block of MASTER,
\begin{equation*}
    \sum_{\tau=t_n}^{E_n} (f^*_\tau - R_\tau)\leq 4 \hat{C}(2^n)+96 \hat{n} \sum_{i=1}^{\ell} \hat{C}(|\mathcal{I}_i|) + 60 \sum_{m=0}^{n} \frac{\rho(2^m)}{\rho(2^n)} \hat{C}(2^m) \log(T/\delta).
\end{equation*}
where $\hat{C}(t)=t\hat{\rho}(t)$, $t_n$ is the start of the block and $E_n\leq t_n+2^n-1$ is the end of the block and  $\mathcal{I}_i$ are consecutive intervals that partition $[t_n,E_n]$ such that $\Delta_{\mathcal{I}_i}\leq \rho(|\mathcal{I}_i|)$ for all $i$.
\label{lemma:lemma_16}
\end{lemma}
Note that the dynamic regret in \citet{wei2021non} is given with respect to $R_t$ rather than its expectation as shown in Definition \ref{def:ns_env}. This is evident in Lemma 3, as  $\sum_{\tau=t_n}^{E_n} (f^*_\tau - R_\tau)$ is upper-bounded with high probability, rather than the expectation $\mathbb{E}[\sum_{\tau=t_n}^{E_n} (f^*_\tau - R_\tau)]$ being upper-bounded. The authors of \citet{wei2021non} choose to express the dynamic regret this way, even though they compare with approaches that follow Definition \ref{def:ns_env} (see e.g. \citet{auer2019adswitch,cheung2020optimism}). However, this can trivially be extended to Definition \ref{def:ns_env}, using the Azuma-Hoeffding inequality, as will be discussed next.

According to Lemma \ref{lemma:lemma_16}, notice that $E_n\leq t_n+2^n-1$, since the block may terminate earlier due to the non-stationarity detection tests being triggered. However, the regret bound is independent of $t_n$ and $E_n$, as it essentially bounds the entire length-$2^n$ block. Thus, let $N_B$ denote the random number of blocks that cover the entire horizon $T$. We denote each block by $B_i$, for all $i\in\{0,1,\dots,N_B-1\}$, where $B_i=[t_i,E_i]$, and $t_0=1,E_{N_{B}-1}=T$. Thus, it is evident that the entire regret over the horizon $T$ can be expressed as follows,
\begin{equation*}
    \sum_{n=0}^{N_B-1}\sum_{\tau=t_n}^{E_n} (f^*_\tau - R_\tau)=\sum_{\tau=1}^{T} (f^*_\tau - R_\tau).
\end{equation*}
Then, by bounding the regret of each different block using Lemma \ref{lemma:lemma_16}, we can provide the following (tight) bound for the regret,
\begin{equation*}
   \sum_{\tau=1}^{T} (f^*_\tau - R_\tau)\leq \sum_{n=0}^{N_B-1} \left[4 \hat{C}(2^n)+96 \hat{n} \sum_{i=1}^{\ell} \hat{C}(|\mathcal{I}_i|) + 60 \sum_{m=0}^{n} \frac{\rho(2^m)}{\rho(2^n)} \hat{C}(2^m) \log(T/\delta)\right].
\end{equation*} 
This reasoning mirrors the approach used in the proof of Theorem 2 from \citet{wei2021non}, as detailed in the Appendices D and E. However, unlike the analysis from \citet{wei2021non}, we do not need to establish an upper bound on the regret in our case. This allows us to bypass the examination of MASTER's behavior between restarts. Instead, summing the relevant terms for the regret, we directly obtain the tightest possible regret bound for MASTER. According to Definition \ref{def:ns_env}, the dynamic regret $\mathrm{Reg}_{\mathrm{D}}(T)$ is given as,
\begin{equation*}
    \mathrm{Reg}_{\mathrm{D}}(T)=\sum_{\tau=1}^{T} (f^*_\tau - R_\tau)+ \sum_{\tau=1}^{T} (R_\tau - \mathbb{E}[R_\tau])
\end{equation*}
which follows from the linearity of the expectation. Hence, this implies that for the dynamic regret of MASTER,
\begin{equation}
   \mathrm{Reg}_{\mathrm{D}}(T)\leq \sum_{n=0}^{N_B-1} \left[4 \hat{C}(2^n)+96 \hat{n} \sum_{i=1}^{\ell} \hat{C}(|\mathcal{I}_i|) + 60 \sum_{m=0}^{n} \frac{\rho(2^m)}{\rho(2^n)} \hat{C}(2^m) \log(T/\delta)\right]+ \sum_{\tau=1}^{T} (R_\tau - \mathbb{E}[R_\tau]).
   \label{eq:append_total_regret}
\end{equation}
Notice that for $\sum_{\tau=1}^{T} (R_\tau - \mathbb{E}[R_\tau])$, we can use the Azuma-Hoeffding inequality to provide a high-probability (with probability at least $1-\delta)$ positive upper bound of sub-linear order in the $\tilde{\mathcal{O}}(\cdot)$ notation, that will not affect the order-optimal regret bounds of MASTER. This extra term, however, is not of interest in our derivation, as we only need to provide a lower-bound on the dynamic regret bound of MASTER, by considering a lower bound on the RHS of Equation \eqref{eq:append_total_regret}. This means that we can only focus on the terms obtained by the regret bound on the blocks of MASTER.

Thus, let $U^M(\mathrm{Reg}_{\mathrm{D}}(T))$ denote the dynamic regret bound of MASTER, that is,
\begin{align*}
    U^M(\mathrm{Reg}_{\mathrm{D}}(T))=\sum_{n=0}^{N_B-1} \left[4 \hat{C}(2^n)+96 \hat{n} \sum_{i=1}^{\ell} \hat{C}(|\mathcal{I}_i|) + 60 \sum_{m=0}^{n} \frac{\rho(2^m)}{\rho(2^n)} \hat{C}(2^m) \log(T/\delta)\right]\\ +\sum_{\tau=1}^{T} (R_\tau - \mathbb{E}[R_\tau]).
\end{align*}

The number of blocks in the horizon $N_B$, depends on the size of the horizon and the non-stationarity detection. It is evident, that the smallest number of blocks is obtained when no restarts happen due to the non-stationarity detection scheme, and in this case $N_B$ becomes deterministic. Thereupon, we can consider the case where no restarts happen due to the non-stationarity detection as a lower bound on $U^M(\mathrm{Reg}_{\mathrm{D}}(T))$. This lower bound essentially corresponds to the case for which the non-stationarity is weak enough for MASTER to handle only with the guarantees of Lemma \ref{lemma_3_master}. However, notice that the following needs to hold,
\begin{equation*}
    \sum_{n=0}^{N_B-1}2^n \geq T.
\end{equation*}
Due to this observation, we can equivalently study the case where we have only a single block with $n=\lceil \log_2 (T)\rceil$. By selecting $n=\lceil \log_2(T)\rceil$ we obtain lower regret compared to the doubling trick that MASTER does. In the case of the doubling trick, the overall performance will deteriorate due to the excess restarting since the horizon is unknown. Conversely, using the knowledge of the horizon will not lead to unnecessary restarts for a given horizon $T$, which leads to a smaller overall regret. Thereupon, studying the case where we fix $n=\lceil \log_2(T)\rceil$, rather than using the doubling trick, is another lower bound for $U^M(\mathrm{Reg}_{\mathrm{D}}(T))$ when we assume no restarts due to the non-stationarity tests. 

Based on the above, we have the following,
\begin{equation*}
U^M(\mathrm{Reg}_{\mathrm{D}}(T))\geq  4 \hat{C}(2^n)+ 60 \frac{\rho(2^n)}{\rho(2^n)} \hat{C}(2^n) \log(T/\delta)
\end{equation*}
as we can only select a single term from the sum $60 \sum_{m=0}^{n} \frac{\rho(2^m)}{\rho(2^n)} \hat{C}(2^m) \log(T/\delta)$ to derive a lower bound on $U^M(\mathrm{Reg}_{\mathrm{D}}(T))$, and since $n=\lceil \log_2(T)\rceil$ and $\hat{C}(t)$ is non-decreasing,
\begin{equation*}
U^M(\mathrm{Reg}_{\mathrm{D}}(T))\geq  4 \hat{C}(T)+ 60 \hat{C}(T) \log(T/\delta).
\end{equation*}
According to the definition of $\hat{C}(t)$ and $\hat{\rho}(t)$, we have that,
\begin{equation*}
    4 \hat{C}(T)+ 60 \hat{C}(T) \log(T/\delta)=24 (\log_2(T) + 1) \log (T/\delta)T \rho(T) \left( 1 + 15 \log(T/\delta) \right).
\end{equation*}
However, based on Assumption \ref{assumption_one}, $\rho(t)\geq 1/\sqrt{t}$ and since $\delta\in(0,1)$,
\begin{equation*}
    24 (\log_2(T) + 1) \log (T/\delta)T \rho(T) \left( 1 + 15 \log(T/\delta) \right)> 24 (\log_2(T) + 1) \log (T)\sqrt{T} \left( 1 + 15 \log(T) \right).
\end{equation*}
Thus, setting,
\begin{equation*}
    B_\mathrm{D}(T)=24 (\log_2(T) + 1) \log (T)\sqrt{T} \left( 1 + 15 \log(T) \right)
\end{equation*}
we finally have that the bound of the dynamic regret of MASTER is lower bounded as,
\begin{equation*}
    U^M(\mathrm{Reg}_{\mathrm{D}}(T))>B_\mathrm{D}(T).
\end{equation*}
According to Definition \ref{def:ns_env}, since the rewards are bounded in $[0,1]$, the worst-case linear dynamic regret is exactly equal to $T$. Moreover, this worst-case regret is a trivial upper bound for the dynamic regret of any algorithm in the NS-RL problem with such rewards, and thus, in order for MASTER's regret bound to not be trivial, we need,
\begin{equation*}
    U^M(\mathrm{Reg}_{\mathrm{D}}(T))<T.
\end{equation*}
However, we can show that for $T\leq 409189687210680$,
\begin{equation*}
    T<B_\mathrm{D}(T)\leq U^M(\mathrm{Reg}_{\mathrm{D}}(T))
\end{equation*}
which means that, whenever
\begin{equation*}
    T\leq 4\times 10^{14},\quad T<U^M(\mathrm{Reg}_{\mathrm{D}}(T))
\end{equation*}
which concludes the proof.

\subsection{Proof Of Theorem \ref{thm:rr_reg}}

We first restate the theorem.

\begin{theorem}
Let the geometric change-point parameter be $\eta = \Theta(T^{-\xi})$, for some $\xi\in(0,1)$. Set the intervals between random restarts of the RR algorithm to be i.i.d. $\mathrm{Geo}(\eta_{\mathrm{R}})$, for some $\eta_{\mathrm{R}}\in (0,1)$ with $\eta_{\mathrm{R}}>\eta$ (order-wise). If the regret of base algorithm is $\mathcal{O}(\mathrm{poly}(\log(T)))$, and $\eta_{\mathrm{R}}=\sqrt{\eta/\mathrm{poly}(\log(T))}$, the RR algorithm is order optimal for PS-MABs.
\end{theorem}
\textbf{Proof}:

To prove the theorem, we first define the notion of (stationary) regret. So far, we have given the definition of dynamic regret as shown in Definition \ref{def:ns_env}. Under a stationiary environment, we have that the reward function is stationary, i.e. $f_1=f_2=...=f_T=f$. Thus, in this case the regret can be defined as,
\begin{equation*}
    \mathrm{Reg}=\mathbb{E}\left[\sum_{t=1}^T (f^* -f(\pi_t)) \right]
\end{equation*}
where $f^* = \max_{\pi\in \Pi} f(\pi)$.
In the case of MABs, the set of policies $\Pi$ is equivalent to the set of arms $\mathcal{A}=\{1,\dots,A\}$. 
Moreover, it is evident that the expectation $f(\pi)$ corresponds to the mean of an arm $\mu_\alpha$, where $\alpha\in\Pi=\mathcal{A}$, i.e.,
\begin{equation*}
    f(\pi)\equiv\mu_a=\mathbb{E}[R_t|\alpha].
\end{equation*}
Thus, the regret of a stationary bandit algorithm $\mathcal{B}$, $R_\mathcal{B}(T)$, essentially corresponds to,
\begin{equation*}
    R_\mathcal{B}=\mathbb{E}\left[\sum_{t=1}^T (\mu_{\alpha^*} -\mu_{A_t})\right]
\end{equation*}
where $\alpha^*=\argmax_{a} \mu_{a}$ and $A_t$ is selected according to the bandit algorithm $\mathcal{B}$. Thus, we can proceed with the analysis of the regret for RR.

Let $(r^{(k)})_{k=1}^\infty$ denote an instance of random restart points drawn such that 
$(r^{(k)}-r^{(k-1)})\stackrel{\text{i.i.d.}}{\sim} \mathrm{Geo}(\eta_{\mathrm{R}})$, with $r^{(0)}:=1$. The total number of the random restarts within the horizon $T$ is given by $N_\mathrm{R}:=\sup\{ k\in\mathbb{N}:r^{(k)}\leq T\}$. Consider a fixed case of the random restart points, where the restart points are set to some arbitrary fixed values in the horizon, expressed by the event $E_{\mathcal{T}}:=\left\{r^{(1)}=t_1,r^{(2)}=t_2,r^{(3)}=t_3,\dots\right\}$, where $\mathcal{T}=\{t_k\}_{k=1}^\infty $ is a fixed, increasing, arbitrary sequence of positive integers, larger than one. Let, $\Delta_{t,a}\:= \max_{a^*} \mu_{t,a^{*}}-\mu_{t,a}$, and since $R_t\in[0,1]$ based on Definition \ref{def:ns_env}, $\Delta_{t,a}\leq 1$ for all $a\in\Pi$ and for all $t$. Thus, the dynamic regret of the RR algorithm can be expressed as
\begin{align}\nonumber
    & \mathrm{Reg}_{\mathrm{D}}(T)=\mathbb{E}\left[\sum_{t=1}^T\Delta_{t,A_t} \right]=\mathbb{E}\left[\sum_{k=1}^{N_\mathrm{R}} \sum_{t=r^{(k-1)}}^{r^{(k)}-1}\Delta_{t,A_t}+\sum_{t=r^{(N_\mathrm{R})}}^T \Delta_{t,A_t} \right].
\end{align}

We focus on the regret accumulated between any two consecutive random restart points, denoted as $r^{(k-1)}=t_{k-1}$ and $r^{(k)}=t_k$. For the purpose of our analysis, we treat these specific instances of restart points as fixed (i.e., we condition on their occurrence), allowing us to ignore the randomness introduced by the random scheduling. Let $C_{k} := \{ \mathrm{no\: change\: in\:} [t_{k - 1}, t_{k} - 1] \}$. Then, we have the following inequality for the regret between consecutive restart points, conditioned on $E_{\mathcal{T}}$,
\begin{align}\nonumber
    \mathbb{E}\left[\sum_{t = r^{(k - 1)}}^{r^{(k)} - 1}\Delta_{t, A_t} \Bigg| E_{\mathcal{T}}\right] &= \Pr \{ C_{k} \} \mathbb{E}\left[\sum_{t = r^{(k - 1)}}^{r^{(k)} - 1}\Delta_{t, A_t} \Bigg| E_{\mathcal{T}}, C_{k} \right] + \Pr \{ \bar{C}_{k} \} \mathbb{E}\left[\sum_{t = r^{(k - 1)}}^{r^{(k)} - 1}\Delta_{t, A_t} \Bigg| E_{\mathcal{T}}, \bar{C}_{k} \right] \\\nonumber
    &\overset{(a)}{\leq} \Pr \{ C_{k} \} R_{\mathcal{B}} (t_{k} - t_{k - 1}) + \Pr \{ \bar{C}_{k} \} (t_{k} - t_{k - 1}) \\\nonumber
    &\overset{(b)}{\leq} R_{\mathcal{B}} (T) + \Pr \{ \bar{C}_{k} \} (t_{k} - t_{k - 1}).
\end{align}
In step $(a)$, we use the crude linear bound $t_{k} - t_{k - 1}$ to bound $\mathbb{E}\left[\sum_{t = r^{(k - 1)}}^{r^{(k)} - 1}\Delta_{t, A_t} \Big| E_{\mathcal{T}}, \bar{C}_{k} \right]$ and $R_{\mathcal{B}} (t_{k} - t_{k - 1})$ to bound $\mathbb{E}\left[\sum_{t = r^{(k - 1)}}^{r^{(k)} - 1}\Delta_{t, A_t} \Big| E_{\mathcal{T}}, C_{k} \right]$, since the piecewise stationary bandit becomes stationary in $[t_{k - 1}, t_{k} - 1]$ given $C_{k}$. In step $(b)$, we apply the fact that $\Pr \{ C_{k} \} \leq 1$ and $R_{\mathcal{B}} (T)$ is increasing with $T$. Then, we derive an upper bound on $\Pr \{ \bar{C}_{k} \}$ as follows:
\begin{align*}
    \Pr \{ \bar{C}_{k} \} &= 1 - \Pr \{ C_{k} \}\\
    &= 1 - (1 - \eta)^{t_{k} - t_{k - 1}} \\
    &\overset{(a)}{\leq} 1 - [1 - \eta (t_{k} - t_{k - 1})] \\
    &= \eta (t_{k} - t_{k - 1})
\end{align*}
where step $(a)$ results from the inequality $(1 - x)^{n} \geq 1 - nx$ for $x \in (0, 1)$ and $n \in \mathbb{N}$. Hence, we have:
\begin{align}\nonumber
    \mathbb{E}\left[\sum_{t = r^{(k - 1)}}^{r^{(k)} - 1}\Delta_{t, A_t} \Bigg| E_{\mathcal{T}}\right] &\leq R_{\mathcal{B}} (T) + \eta (t_{k} - t_{k - 1})^{2}.
\end{align}

Following the previous reasoning, we equivalently have that,
\begin{align*}
     & \mathbb{E}\left[\sum_{t=t_{N_\mathrm{R}}}^{T}\Delta_{t,A_t}\Bigg| E_{\mathcal{T}}\right]\leq R_{\mathcal{B}} (T) + \eta (t_{N_{\mathrm{R}}+1} - t_{N_{\mathrm{R}}})^{2}.
\end{align*}
Thereupon, since the previous bounds hold for any arbitrary set of random restarting points, we can substitute the random variables and upper bound the initial regret as follows:
\begin{flalign}\nonumber
    \mathrm{Reg}_{\mathrm{D}}(T)&\leq \mathbb{E}\Biggl[\sum_{k=1}^{N_\mathrm{R}}\left[ R_{\mathcal{B}} (T) + \eta \left(r^{(k)} - r^{(k - 1)}\right)^{2} \right] + R_{\mathcal{B}} (T) + \eta \left(r^{(N_{\mathrm{R}} + 1)} - r^{(N_{\mathrm{R}})}\right)^{2} \Biggr]\\\nonumber
    & = \mathbb{E}[N_\mathrm{R}+1] R_{\mathcal{B}}\left(T\right) + \eta \mathbb{E}\left[\sum_{k=1}^{N_\mathrm{R}+1}\left(r^{(k)}-r^{(k-1)}\right)^2\right]\\\nonumber 
    & = \mathbb{E}[N_\mathrm{R}+1] R_{\mathcal{B}}\left(T\right) +\eta \mathbb{E}[N_\mathrm{R}+1]\mathbb{E}\left[\left(r^{(1)}-r^{(0)}\right)^2\right]\\\nonumber 
    & = (\eta_{\mathrm{R}} T+1)R_{\mathcal{B}}\left(T\right)+\eta (\eta_{\mathrm{R}} T +1)\left( \frac{2-\eta_{\mathrm{R}}}{\eta_{\mathrm{R}}^2}\right) \nonumber
\end{flalign}
where the above is derived using the geometric i.i.d. assumption and Wald's Identity, as $N_\mathrm{R}$ is a stopping time. Hence, since $R_{\mathcal{B}}\left(T\right) =\mathcal{O}(\mathrm{poly}(\log(T)))$, the following holds,
\begin{align}\nonumber
    &\mathrm{Reg}_{\mathrm{D}}(T)\leq \frac{2\eta T}{\eta_{\mathrm{R}}}+\eta_{\mathrm{R}} T R_{\mathcal{B}}\left(T\right)+\eta\left( \frac{2-\eta_{\mathrm{R}}}{\eta_{\mathrm{R}}^2}\right) +R_{\mathcal{B}}\left(T\right)\\\nonumber
    &\mathrm{Reg}_{\mathrm{D}}(T)=\mathcal{O}\left( \frac{2\eta T}{\eta_{\mathrm{R}}}+\eta_{\mathrm{R}} T\mathrm{poly}(\log(T))+\eta\left( \frac{2-\eta_{\mathrm{R}}}{\eta_{\mathrm{R}}^2}\right)+\mathrm{poly}(\log(T))\right).
\end{align}
Ignoring the constants that are independent of the horizon, since $\eta_{\mathrm{R}}>\eta$ (order-wise) and $\eta_{\mathrm{R}}\in(0,1)$, it is evident that the leading terms in the regret, correspond to $\frac{\eta T}{\eta_{\mathrm{R}}}$ and $\eta_{\mathrm{R}} T\mathrm{poly}(\log(T))$, and in order to ensure that we have the appropriate order,
\begin{align}\nonumber
    &\eta T/\eta_{\mathrm{R}} = \eta_{\mathrm{R}} T\mathrm{poly}(\log(T)) \Rightarrow \eta_{\mathrm{R}}=\sqrt{\eta/\mathrm{poly}(\log(T))}.
\end{align}
Thus, we have that, 
\begin{align}\nonumber
    \mathrm{Reg}_{\mathrm{D}}(T)&=\mathcal{O} \Biggr( \frac{2 \eta T \sqrt{\mathrm{poly}(\log(T))}}{\sqrt{\eta}}+\frac{\sqrt{\eta}}{\sqrt{\mathrm{poly}(\log(T))}} T\mathrm{poly}(\log(T))\\\nonumber
    &+\eta\left( \frac{2\mathrm{poly}(\log(T))-\sqrt{\eta \mathrm{poly}(\log(T))}}{\eta}\right)+\mathrm{poly}(\log(T))\Biggr)\\\nonumber
    &=\mathcal{O}\left(3\sqrt{\eta}T\sqrt{\mathrm{poly}(\log(T))} -\sqrt{\eta \mathrm{poly}(\log(T))} +3\mathrm{poly}(\log(T))\right).\nonumber
\end{align}
Finally, since $\eta=\Theta(T^{-\xi}),\:\xi\in(0,1)$, we have that,
\begin{align}\nonumber
    \mathrm{Reg}_{\mathrm{D}}(T)&= \mathcal{O}\left(3T^{1-\xi/2}\sqrt{\mathrm{poly}(\log(T))} -T^{-\xi/2}\sqrt{ \mathrm{poly}(\log(T))} +3\mathrm{poly}(\log(T))\right)\nonumber
\end{align}
which implies,
\begin{align}\nonumber    
     \mathrm{Reg}_{\mathrm{D}}(T)&= \mathcal{O} \left(3T^{1-\xi/2}\sqrt{\mathrm{poly}(\log(T))} +3\mathrm{poly}(\log(T))\right)\nonumber.
\end{align}
Hence, it is evident that, $\mathrm{Reg}_{\mathrm{D}}(T)= \tilde{\mathcal{O}}(T^{1-\xi/2})$. 

To show that the RR algorithm is order-optimal, we need to show that the optimal minimax bound is $\tilde{\Omega}(T^{1-\xi/2})$ as well.

In the context of NS-MABs, the non-stationarity measure, according to \citet{wei2021non}, is defined as $\Delta(t) = \max_{\alpha} |\mu_{t,\alpha} - \mu_{t+1,\alpha}|$. In the PS-MAB setting, each change-point introduces a non-negative non-stationarity $\Delta(t)$. Hence, the number of changes $L=1+\sum_{\tau=1}^{T-1}\mathds{1}{[\Delta(\tau)\neq 0]}$ and the cumulative non-stationarity $\Delta=\sum_{\tau=1}^{T-1} \Delta(\tau)$ have the same order in $T$. Therefore, the minimax optimal regret bound in $\tilde{\mathcal{O}}(\cdot)$ notation is $\min\left\{ \Delta^{1/3}T^{2/3} + \sqrt{T}, \sqrt{LT} \right\} = \sqrt{LT}$. To demonstrate that RR is order-optimal, we analyze the behavior of $\sqrt{LT}$.

Due to the geometric change-point assumption, it is evident that the number of changes $L$ is distributed as a binomial random variable, i.e., $L\sim \mathrm{B}(T,\eta)$. Now, first notice that by Jensen's inequality,
\begin{equation*}
    \mathbb{E}[\sqrt{L}]\leq \sqrt{\mathbb{E}[L]}
\end{equation*}
and by the definition of the binomial random variable, $\mathbb{E}[L]=\eta T$, which implies,
\begin{equation*}
    \mathbb{E}[\sqrt{L}]\leq \sqrt{\eta T}.
\end{equation*}
On the other hand, since $\sqrt{x} \geq 1 + \frac{x - 1}{2} - \frac{(x - 1)^{2}}{2}$ for any $x \geq 0$, by plugging $\frac{L}{\mathbb{E}[L]}$ into $x$ and then taking expectation, we have,
\begin{align*}
     \frac{\mathbb{E}[\sqrt{L}]}{\sqrt{\mathbb{E}[L]}} = \mathbb{E} \left[ \sqrt{\frac{L}{\mathbb{E}[L]}} \right] \geq \mathbb{E} \left[1 + \frac{L/\mathbb{E}[L] - 1}{2} - \frac{( L/\mathbb{E}[L] - 1 )^{2}}{2} \right] = 1 - \frac{\mathrm{Var}(L)}{2(\mathbb{E}[L])^{2}}.
\end{align*}
Since $\mathbb{E}[L] = \eta T$ and $\mathrm{Var}(L) = \eta (1-\eta) T$, we have,
\begin{align*}
    \mathbb{E}[\sqrt{L}] \geq \sqrt{\eta T} \left( 1 - \frac{ 1 - \eta}{2 \eta T} \right).
\end{align*}
Because $\eta=\Theta (T^{-\xi})$ and $\xi\in(0,1)$, we can show that 
\begin{align*}
    &\mathbb{E} \left[\sqrt{LT}\right] = \tilde{\Omega} \left( T^{(2 - \xi)/2} \left( 1 - \frac{1 - T^{- \xi}}{2 T^{(1 - \xi)}} \right) \right) = \tilde{\Omega} \left( T^{1 - \xi/2} \right).
\end{align*}
Hence, the RR algorithm is order-optimal. Note that RR can be made order-optimal for a case of deterministic change-points, as long as we focus on an instance of the geometric change-point model.

\section{Experimental Details}
\subsection{Technical Details}
The experiments were conducted using the C++ programming language. Moreover, all the experiments were conducted on a server with an Intel Xeon Gold 6148 CPU and allocating 200 GB of memory. 
\subsection{Environment Details}
\paragraph{Geometric Change-Points.} 
The change-points are generated using a geometric distribution. More specifically, for a given horizon $T$, the probability of the geometric distribution is set exactly equal to $\eta=T^{-\xi}$ for some $\xi\in(0,1)$.
\paragraph{Deterministic Change-Points.} 
For a given horizon $T$, the number of change-points is set equal to $N_C$, with $N_C<T$. Then, the change-points are placed every $\mathrm{round}(T/N_C)$ time-steps in the horizon, where $\mathrm{round}(\cdot)$ corresponds to the rounding operation. This ensures that the change-points are roughly evenly placed. In order to ensure that RR can be applied in this deterministic scenario, we fix a geometric parameter $\eta$ and we set the number of change-points equal to $N_C=\lceil \eta T \rceil$ in all of the runs.

\paragraph{Uniform Problem.} 
For this problem, the means of all arms are initialized in $[0,1]$. Then in each change-point, the number of arms that are going to change is chosen uniformly, but is always greater than one. For each arm, the magnitude of the change in uniformly selected in $[0.1,0.4]$ with probability 0.5 to be a positive or negative change. If a change results in a mean exceeding 1, the change is subtracted instead. Similarly, if it would result in a mean below 0, the change is added.

\paragraph{Wost-Case Problem.} 
In this problem, the means of all arms are initialized at 0.3, plus a small positive offset uniformly selected from $[0.0005,0.005]$, ensuring that the means remain close to each other. At each change point, the arm with the lowest mean changes to match the highest mean plus a small gap uniformly selected from  $[0.005,0.05]$, ensuring that the change is small, but enough to affect the environment. If the mean of any arm exceeds 0.99 at a change point, then the process is reset and follows the same initialization, but ensuring that the arm with the lowest mean, becomes the arm with the highest mean.

\subsection{Algorithm Details}
The algorithm of GLRklUCB is employed with global restarts and the Bernoulli GLR change-point detector, as specified in \citet{besson2022efficient}. The Bernoulli GLR change-point detector is also used for the QCD+ algorithm. When employing the Bernoulli GLR change-point detector, as it was similarly suggested in \citet{besson2022efficient}, its threshold is set to be equal to $\beta(n,\delta)=\log (4n\sqrt{n}/\delta)$.

\subsection{Additional Results}
The experiments were conducted for horizons ranging from $T=1000$ to $T=100000$, with increments of approximately $5000$, averaged over $4000$ independent trials. For geometric change-points, we use $\xi \in \{0.3, 0.4, 0.5, 0.6, 0.7, 0.8\}$, and for deterministic change-points, $N_C \in \{\lceil T^{0.7} \rceil, \lceil T^{0.6} \rceil, \lceil T^{0.5} \rceil, \lceil T^{0.4} \rceil, \lceil T^{0.3} \rceil, \lceil T^{0.2} \rceil\}$. 

Due to minimal differences observed at every $5000$-step increment, we present the using horizon increments of $10000$, with the addition of $T=1000,T=2000$ and $T=5000$. 

Moreover, since the experiments were conducted using the C++ programming language, which is very fast in computations, there was no significant difference in the average running time until the horizon became adequately large among the methods, with the exception of QCD+UCB, which was always significantly faster. Given the significant computational demands of MASTER, the lack of a noticeable difference for these horizons is due to the complexity of the klUCB algorithm, which is computationally demanding. Therefore, we only report the average running time for horizons of $T=60000$ and beyond, while the dynamic regret and non-stationarity detection results are presented for the full range mentioned earlier.
\begin{table}[H]
\caption{Average Time Per Run (sec) For Different Change-Points, $T$=60000, 4000 Runs.}
\label{table:time_per_exp_T60000}
\tiny
\setlength{\tabcolsep}{3pt} 
\begin{center}
\begin{tabular}{c|c|cccccc}
\multicolumn{8}{c}{\textbf{(a) Random Change-Points (Columns: $\xi$)}} \\ \hline
\multicolumn{1}{c|}{\bf Algorithm} & \multicolumn{1}{c|}{\bf PB} & \multicolumn{1}{c}{\bf 0.3} & \multicolumn{1}{c}{\bf 0.4} & \multicolumn{1}{c}{\bf 0.5} & \multicolumn{1}{c}{\bf 0.6} & \multicolumn{1}{c}{\bf 0.7} & \multicolumn{1}{c}{\bf 0.8} \\
\hline
{MASTER} & {Unif} & {1.39 $\pm$ 0.00} & {1.39 $\pm$ 0.00} & {1.39 $\pm$ 0.00} & {1.39 $\pm$ 0.00} & {1.40 $\pm$ 0.06} & {1.40 $\pm$ 0.04} \\
{MASTER} & {Worst} & {1.39 $\pm$ 0.01} & {1.39 $\pm$ 0.01} & {1.39 $\pm$ 0.00} & {1.39 $\pm$ 0.04} & {1.39 $\pm$ 0.04} & {1.39 $\pm$ 0.04} \\
{RR\_p.05} & {Unif} & {0.68 $\pm$ 0.01} & {0.68 $\pm$ 0.01} & {0.67 $\pm$ 0.01} & {0.67 $\pm$ 0.02} & {0.67 $\pm$ 0.03} & {0.67 $\pm$ 0.04} \\
{RR\_p.05} & {Worst} & {0.58 $\pm$ 0.01} & {0.57 $\pm$ 0.02} & {0.57 $\pm$ 0.02} & {0.58 $\pm$ 0.03} & {0.64 $\pm$ 0.05} & {0.74 $\pm$ 0.02} \\
{GLRklUCB} & {Unif} & {0.83 $\pm$ 0.01} & {0.82 $\pm$ 0.01} & {0.86 $\pm$ 0.01} & {0.95 $\pm$ 0.03} & {1.17 $\pm$ 0.09} & {1.75 $\pm$ 0.41} \\
{GLRklUCB} & {Worst} & {0.79 $\pm$ 0.01} & {0.88 $\pm$ 0.02} & {0.97 $\pm$ 0.05} & {1.18 $\pm$ 0.10} & {1.66 $\pm$ 0.23} & {2.68 $\pm$ 0.78} \\
{QCD+UCB} & {Unif} & {0.05 $\pm$ 0.01} & {0.05 $\pm$ 0.00} & {0.06 $\pm$ 0.01} & {0.11 $\pm$ 0.02} & {0.30 $\pm$ 0.10} & {0.86 $\pm$ 0.42} \\
{QCD+UCB} & {Worst} & {0.04 $\pm$ 0.00} & {0.06 $\pm$ 0.01} & {0.09 $\pm$ 0.01} & {0.17 $\pm$ 0.04} & {0.34 $\pm$ 0.10} & {0.70 $\pm$ 0.30} \\
{QCD+klUCB} & {Unif} & {1.01 $\pm$ 0.00} & {1.01 $\pm$ 0.01} & {1.03 $\pm$ 0.01} & {1.10 $\pm$ 0.03} & {1.32 $\pm$ 0.10} & {1.93 $\pm$ 0.44} \\
{QCD+klUCB} & {Worst} & {1.00 $\pm$ 0.01} & {1.06 $\pm$ 0.04} & {1.17 $\pm$ 0.13} & {1.42 $\pm$ 0.32} & {1.79 $\pm$ 0.24} & {2.83 $\pm$ 0.80} \\
\hline
\end{tabular}

\vspace{0.5em} 

\begin{tabular}{c|c|cccccc}
\multicolumn{8}{c}{\textbf{(b) Deterministic Change-Points (Columns: $N_C$)}} \\ \hline
\multicolumn{1}{c|}{\bf Algorithm} & \multicolumn{1}{c|}{\bf PB} & \multicolumn{1}{c}{\bf 2212} & \multicolumn{1}{c}{\bf 737} & \multicolumn{1}{c}{\bf 245} & \multicolumn{1}{c}{\bf 82} & \multicolumn{1}{c}{\bf 28} & \multicolumn{1}{c}{\bf 10} \\
\hline
{MASTER} & {Unif} & {1.40 $\pm$ 0.00} & {1.41 $\pm$ 0.07} & {1.40 $\pm$ 0.01} & {1.40 $\pm$ 0.01} & {1.41 $\pm$ 0.19} & {1.43 $\pm$ 0.16} \\
{MASTER} & {Worst} & {1.40 $\pm$ 0.01} & {1.40 $\pm$ 0.05} & {1.41 $\pm$ 0.11} & {1.41 $\pm$ 0.01} & {1.40 $\pm$ 0.02} & {1.42 $\pm$ 0.20} \\
{RR\_p.05} & {Unif} & {0.68 $\pm$ 0.00} & {0.68 $\pm$ 0.05} & {0.67 $\pm$ 0.00} & {0.65 $\pm$ 0.01} & {0.72 $\pm$ 0.12} & {0.70 $\pm$ 0.11} \\
{RR\_p.05} & {Worst} & {0.58 $\pm$ 0.01} & {0.58 $\pm$ 0.01} & {0.59 $\pm$ 0.05} & {0.62 $\pm$ 0.01} & {0.60 $\pm$ 0.00} & {0.77 $\pm$ 0.14} \\
{GLRklUCB} & {Unif} & {0.87 $\pm$ 0.01} & {0.84 $\pm$ 0.05} & {0.84 $\pm$ 0.01} & {0.89 $\pm$ 0.02} & {1.01 $\pm$ 0.15} & {1.32 $\pm$ 0.18} \\
{GLRklUCB} & {Worst} & {0.79 $\pm$ 0.01} & {0.89 $\pm$ 0.03} & {0.98 $\pm$ 0.08} & {1.15 $\pm$ 0.06} & {1.43 $\pm$ 0.06} & {2.07 $\pm$ 0.29} \\
{QCD+UCB} & {Unif} & {0.07 $\pm$ 0.01} & {0.05 $\pm$ 0.02} & {0.05 $\pm$ 0.00} & {0.07 $\pm$ 0.00} & {0.15 $\pm$ 0.06} & {0.42 $\pm$ 0.10} \\
{QCD+UCB} & {Worst} & {0.05 $\pm$ 0.00} & {0.07 $\pm$ 0.03} & {0.10 $\pm$ 0.03} & {0.17 $\pm$ 0.00} & {0.21 $\pm$ 0.00} & {0.52 $\pm$ 0.11} \\
{QCD+klUCB} & {Unif} & {1.02 $\pm$ 0.01} & {1.02 $\pm$ 0.06} & {1.02 $\pm$ 0.00} & {1.03 $\pm$ 0.01} & {1.14 $\pm$ 0.16} & {1.43 $\pm$ 0.16} \\
{QCD+klUCB} & {Worst} & {1.00 $\pm$ 0.00} & {1.06 $\pm$ 0.02} & {1.42 $\pm$ 0.28} & {1.24 $\pm$ 0.02} & {1.58 $\pm$ 0.08} & {2.22 $\pm$ 0.34} \\
\hline
\end{tabular}

\end{center}
\end{table}

\begin{table}[H]
\caption{Average Time Per Run (sec) For Different Change-Points, $T$=70000, 4000 Runs.}
\label{table:time_per_exp_T70000}
\tiny
\setlength{\tabcolsep}{3pt} 
\begin{center}
\begin{tabular}{c|c|cccccc}
\multicolumn{8}{c}{\textbf{(a) Random Change-Points (Columns: $\xi$)}} \\ \hline
\multicolumn{1}{c|}{\bf Algorithm} & \multicolumn{1}{c|}{\bf PB} & \multicolumn{1}{c}{\bf 0.3} & \multicolumn{1}{c}{\bf 0.4} & \multicolumn{1}{c}{\bf 0.5} & \multicolumn{1}{c}{\bf 0.6} & \multicolumn{1}{c}{\bf 0.7} & \multicolumn{1}{c}{\bf 0.8} \\
\hline
{MASTER} & {Unif} & {1.76 $\pm$ 0.01} & {1.76 $\pm$ 0.00} & {1.76 $\pm$ 0.01} & {1.76 $\pm$ 0.01} & {1.76 $\pm$ 0.07} & {1.77 $\pm$ 0.13} \\
{MASTER} & {Worst} & {1.76 $\pm$ 0.00} & {1.76 $\pm$ 0.02} & {1.76 $\pm$ 0.02} & {1.76 $\pm$ 0.01} & {1.76 $\pm$ 0.01} & {1.76 $\pm$ 0.02} \\
{RR\_p.05} & {Unif} & {0.79 $\pm$ 0.01} & {0.79 $\pm$ 0.01} & {0.79 $\pm$ 0.01} & {0.78 $\pm$ 0.02} & {0.79 $\pm$ 0.05} & {0.79 $\pm$ 0.05} \\
{RR\_p.05} & {Worst} & {0.67 $\pm$ 0.01} & {0.66 $\pm$ 0.01} & {0.66 $\pm$ 0.03} & {0.68 $\pm$ 0.04} & {0.74 $\pm$ 0.06} & {0.86 $\pm$ 0.03} \\
{GLRklUCB} & {Unif} & {0.97 $\pm$ 0.01} & {0.96 $\pm$ 0.01} & {1.00 $\pm$ 0.01} & {1.12 $\pm$ 0.03} & {1.40 $\pm$ 0.11} & {2.19 $\pm$ 0.52} \\
{GLRklUCB} & {Worst} & {0.92 $\pm$ 0.01} & {1.03 $\pm$ 0.03} & {1.15 $\pm$ 0.07} & {1.40 $\pm$ 0.17} & {2.03 $\pm$ 0.30} & {3.44 $\pm$ 1.08} \\
{QCD+UCB} & {Unif} & {0.06 $\pm$ 0.01} & {0.06 $\pm$ 0.00} & {0.07 $\pm$ 0.01} & {0.14 $\pm$ 0.02} & {0.39 $\pm$ 0.12} & {1.14 $\pm$ 0.53} \\
{QCD+UCB} & {Worst} & {0.05 $\pm$ 0.00} & {0.07 $\pm$ 0.01} & {0.10 $\pm$ 0.01} & {0.20 $\pm$ 0.04} & {0.42 $\pm$ 0.13} & {0.91 $\pm$ 0.40} \\
{QCD+klUCB} & {Unif} & {1.18 $\pm$ 0.00} & {1.18 $\pm$ 0.01} & {1.20 $\pm$ 0.01} & {1.30 $\pm$ 0.03} & {1.58 $\pm$ 0.12} & {2.41 $\pm$ 0.57} \\
{QCD+klUCB} & {Worst} & {1.16 $\pm$ 0.01} & {1.25 $\pm$ 0.05} & {1.39 $\pm$ 0.17} & {1.73 $\pm$ 0.46} & {2.19 $\pm$ 0.30} & {3.66 $\pm$ 1.13} \\
\hline
\end{tabular}

\vspace{0.5em} 

\begin{tabular}{c|c|cccccc}
\multicolumn{8}{c}{\textbf{(b) Deterministic Change-Points (Columns: $N_C$)}} \\ \hline
\multicolumn{1}{c|}{\bf Algorithm} & \multicolumn{1}{c|}{\bf PB} & \multicolumn{1}{c}{\bf 2464} & \multicolumn{1}{c}{\bf 808} & \multicolumn{1}{c}{\bf 265} & \multicolumn{1}{c}{\bf 87} & \multicolumn{1}{c}{\bf 29} & \multicolumn{1}{c}{\bf 10} \\
\hline
{MASTER} & {Unif} & {1.76 $\pm$ 0.01} & {1.83 $\pm$ 0.02} & {1.80 $\pm$ 0.01} & {1.77 $\pm$ 0.00} & {1.79 $\pm$ 0.11} & {1.85 $\pm$ 0.20} \\
{MASTER} & {Worst} & {1.77 $\pm$ 0.01} & {1.79 $\pm$ 0.06} & {1.81 $\pm$ 0.12} & {1.81 $\pm$ 0.11} & {1.77 $\pm$ 0.08} & {1.82 $\pm$ 0.09} \\
{RR\_p.05} & {Unif} & {0.79 $\pm$ 0.00} & {0.79 $\pm$ 0.01} & {0.79 $\pm$ 0.00} & {0.80 $\pm$ 0.00} & {0.82 $\pm$ 0.05} & {0.75 $\pm$ 0.19} \\
{RR\_p.05} & {Worst} & {0.67 $\pm$ 0.01} & {0.66 $\pm$ 0.03} & {0.68 $\pm$ 0.01} & {0.69 $\pm$ 0.05} & {0.71 $\pm$ 0.03} & {0.86 $\pm$ 0.03} \\
{GLRklUCB} & {Unif} & {1.02 $\pm$ 0.01} & {0.96 $\pm$ 0.01} & {0.97 $\pm$ 0.01} & {1.05 $\pm$ 0.05} & {1.20 $\pm$ 0.07} & {1.58 $\pm$ 0.13} \\
{GLRklUCB} & {Worst} & {0.93 $\pm$ 0.01} & {1.09 $\pm$ 0.03} & {1.15 $\pm$ 0.07} & {1.35 $\pm$ 0.11} & {1.95 $\pm$ 0.11} & {2.88 $\pm$ 0.36} \\
{QCD+UCB} & {Unif} & {0.11 $\pm$ 0.00} & {0.06 $\pm$ 0.00} & {0.06 $\pm$ 0.00} & {0.09 $\pm$ 0.00} & {0.18 $\pm$ 0.03} & {0.52 $\pm$ 0.13} \\
{QCD+UCB} & {Worst} & {0.07 $\pm$ 0.00} & {0.07 $\pm$ 0.01} & {0.11 $\pm$ 0.02} & {0.19 $\pm$ 0.04} & {0.34 $\pm$ 0.04} & {0.48 $\pm$ 0.01} \\
{QCD+klUCB} & {Unif} & {1.19 $\pm$ 0.01} & {1.18 $\pm$ 0.02} & {1.18 $\pm$ 0.00} & {1.24 $\pm$ 0.05} & {1.35 $\pm$ 0.06} & {1.73 $\pm$ 0.21} \\
{QCD+klUCB} & {Worst} & {1.17 $\pm$ 0.01} & {1.35 $\pm$ 0.04} & {1.46 $\pm$ 0.10} & {1.52 $\pm$ 0.12} & {2.05 $\pm$ 0.09} & {3.02 $\pm$ 0.38} \\
\hline
\end{tabular}

\end{center}
\end{table}

\begin{table}[H]
\caption{Average Time Per Run (sec) For Different Change-Points, $T$=80000, 4000 Runs.}
\label{table:time_per_exp_T80000}
\tiny
\setlength{\tabcolsep}{3pt} 
\begin{center}
\begin{tabular}{c|c|cccccc}
\multicolumn{8}{c}{\textbf{(a) Random Change-Points (Columns: $\xi$)}} \\ \hline
\multicolumn{1}{c|}{\bf Algorithm} & \multicolumn{1}{c|}{\bf PB} & \multicolumn{1}{c}{\bf 0.3} & \multicolumn{1}{c}{\bf 0.4} & \multicolumn{1}{c}{\bf 0.5} & \multicolumn{1}{c}{\bf 0.6} & \multicolumn{1}{c}{\bf 0.7} & \multicolumn{1}{c}{\bf 0.8} \\
\hline
{MASTER} & {Unif} & {2.06 $\pm$ 0.01} & {2.06 $\pm$ 0.00} & {2.06 $\pm$ 0.00} & {2.07 $\pm$ 0.01} & {2.07 $\pm$ 0.11} & {2.07 $\pm$ 0.11} \\
{MASTER} & {Worst} & {2.07 $\pm$ 0.02} & {2.06 $\pm$ 0.02} & {2.06 $\pm$ 0.03} & {2.06 $\pm$ 0.06} & {2.06 $\pm$ 0.06} & {2.07 $\pm$ 0.10} \\
{RR\_p.05} & {Unif} & {0.90 $\pm$ 0.01} & {0.90 $\pm$ 0.01} & {0.90 $\pm$ 0.02} & {0.90 $\pm$ 0.03} & {0.90 $\pm$ 0.06} & {0.90 $\pm$ 0.06} \\
{RR\_p.05} & {Worst} & {0.77 $\pm$ 0.01} & {0.76 $\pm$ 0.01} & {0.76 $\pm$ 0.02} & {0.77 $\pm$ 0.04} & {0.83 $\pm$ 0.06} & {0.98 $\pm$ 0.04} \\
{GLRklUCB} & {Unif} & {1.11 $\pm$ 0.01} & {1.10 $\pm$ 0.01} & {1.15 $\pm$ 0.02} & {1.29 $\pm$ 0.03} & {1.69 $\pm$ 0.16} & {2.68 $\pm$ 0.68} \\
{GLRklUCB} & {Worst} & {1.06 $\pm$ 0.01} & {1.18 $\pm$ 0.03} & {1.33 $\pm$ 0.09} & {1.66 $\pm$ 0.33} & {2.43 $\pm$ 0.38} & {4.24 $\pm$ 1.34} \\
{QCD+UCB} & {Unif} & {0.07 $\pm$ 0.01} & {0.06 $\pm$ 0.00} & {0.09 $\pm$ 0.01} & {0.18 $\pm$ 0.03} & {0.49 $\pm$ 0.14} & {1.46 $\pm$ 0.69} \\
{QCD+UCB} & {Worst} & {0.06 $\pm$ 0.01} & {0.08 $\pm$ 0.01} & {0.12 $\pm$ 0.02} & {0.25 $\pm$ 0.05} & {0.50 $\pm$ 0.15} & {1.14 $\pm$ 0.50} \\
{QCD+klUCB} & {Unif} & {1.34 $\pm$ 0.01} & {1.35 $\pm$ 0.01} & {1.38 $\pm$ 0.01} & {1.49 $\pm$ 0.04} & {1.85 $\pm$ 0.15} & {2.92 $\pm$ 0.72} \\
{QCD+klUCB} & {Worst} & {1.33 $\pm$ 0.02} & {1.43 $\pm$ 0.05} & {1.62 $\pm$ 0.19} & {2.04 $\pm$ 0.58} & {2.63 $\pm$ 0.38} & {4.50 $\pm$ 1.39} \\
\hline
\end{tabular}

\vspace{0.5em} 

\begin{tabular}{c|c|cccccc}
\multicolumn{8}{c}{\textbf{(b) Deterministic Change-Points (Columns: $N_C$)}} \\ \hline
\multicolumn{1}{c|}{\bf Algorithm} & \multicolumn{1}{c|}{\bf PB} & \multicolumn{1}{c}{\bf 2705} & \multicolumn{1}{c}{\bf 875} & \multicolumn{1}{c}{\bf 283} & \multicolumn{1}{c}{\bf 92} & \multicolumn{1}{c}{\bf 30} & \multicolumn{1}{c}{\bf 10} \\
\hline
{MASTER} & {Unif} & {2.07 $\pm$ 0.01} & {2.13 $\pm$ 0.10} & {2.13 $\pm$ 0.13} & {2.11 $\pm$ 0.04} & {2.07 $\pm$ 0.01} & {2.12 $\pm$ 0.06} \\
{MASTER} & {Worst} & {2.10 $\pm$ 0.06} & {2.10 $\pm$ 0.05} & {2.09 $\pm$ 0.04} & {2.07 $\pm$ 0.02} & {2.13 $\pm$ 0.03} & {2.09 $\pm$ 0.04} \\
{RR\_p.05} & {Unif} & {0.90 $\pm$ 0.02} & {0.90 $\pm$ 0.03} & {0.90 $\pm$ 0.04} & {0.93 $\pm$ 0.02} & {0.89 $\pm$ 0.01} & {0.89 $\pm$ 0.02} \\
{RR\_p.05} & {Worst} & {0.78 $\pm$ 0.04} & {0.75 $\pm$ 0.04} & {0.73 $\pm$ 0.03} & {0.74 $\pm$ 0.01} & {0.83 $\pm$ 0.01} & {0.97 $\pm$ 0.03} \\
{GLRklUCB} & {Unif} & {1.16 $\pm$ 0.01} & {1.11 $\pm$ 0.10} & {1.14 $\pm$ 0.08} & {1.22 $\pm$ 0.02} & {1.41 $\pm$ 0.01} & {1.73 $\pm$ 0.06} \\
{GLRklUCB} & {Worst} & {1.06 $\pm$ 0.06} & {1.22 $\pm$ 0.05} & {1.35 $\pm$ 0.19} & {1.66 $\pm$ 0.29} & {2.35 $\pm$ 0.18} & {3.22 $\pm$ 0.37} \\
{QCD+UCB} & {Unif} & {0.10 $\pm$ 0.00} & {0.06 $\pm$ 0.01} & {0.07 $\pm$ 0.02} & {0.10 $\pm$ 0.01} & {0.23 $\pm$ 0.02} & {0.48 $\pm$ 0.01} \\
{QCD+UCB} & {Worst} & {0.06 $\pm$ 0.01} & {0.09 $\pm$ 0.02} & {0.10 $\pm$ 0.01} & {0.19 $\pm$ 0.01} & {0.45 $\pm$ 0.01} & {0.61 $\pm$ 0.03} \\
{QCD+klUCB} & {Unif} & {1.36 $\pm$ 0.01} & {1.36 $\pm$ 0.03} & {1.37 $\pm$ 0.07} & {1.43 $\pm$ 0.03} & {1.58 $\pm$ 0.01} & {1.91 $\pm$ 0.02} \\
{QCD+klUCB} & {Worst} & {1.33 $\pm$ 0.05} & {1.47 $\pm$ 0.05} & {1.70 $\pm$ 0.15} & {1.89 $\pm$ 0.31} & {2.48 $\pm$ 0.18} & {3.38 $\pm$ 0.39} \\
\hline
\end{tabular}

\end{center}
\end{table}

\begin{table}[H]
\caption{Average Time Per Run (sec) For Different Change-Points, $T$=90000, 4000 Runs.}
\label{table:time_per_exp_T90000}
\tiny
\setlength{\tabcolsep}{3pt} 
\begin{center}
\begin{tabular}{c|c|cccccc}
\multicolumn{8}{c}{\textbf{(a) Random Change-Points (Columns: $\xi$)}} \\ \hline
\multicolumn{1}{c|}{\bf Algorithm} & \multicolumn{1}{c|}{\bf PB} & \multicolumn{1}{c}{\bf 0.3} & \multicolumn{1}{c}{\bf 0.4} & \multicolumn{1}{c}{\bf 0.5} & \multicolumn{1}{c}{\bf 0.6} & \multicolumn{1}{c}{\bf 0.7} & \multicolumn{1}{c}{\bf 0.8} \\
\hline
{MASTER} & {Unif} & {2.58 $\pm$ 0.02} & {2.58 $\pm$ 0.02} & {2.58 $\pm$ 0.00} & {2.58 $\pm$ 0.05} & {2.58 $\pm$ 0.06} & {2.60 $\pm$ 0.19} \\
{MASTER} & {Worst} & {2.58 $\pm$ 0.01} & {2.58 $\pm$ 0.00} & {2.58 $\pm$ 0.03} & {2.58 $\pm$ 0.04} & {2.59 $\pm$ 0.11} & {2.59 $\pm$ 0.13} \\
{RR\_p.05} & {Unif} & {1.02 $\pm$ 0.01} & {1.01 $\pm$ 0.01} & {1.01 $\pm$ 0.02} & {1.01 $\pm$ 0.05} & {1.01 $\pm$ 0.08} & {1.02 $\pm$ 0.10} \\
{RR\_p.05} & {Worst} & {0.87 $\pm$ 0.01} & {0.85 $\pm$ 0.02} & {0.85 $\pm$ 0.03} & {0.87 $\pm$ 0.06} & {0.93 $\pm$ 0.08} & {1.10 $\pm$ 0.02} \\
{GLRklUCB} & {Unif} & {1.24 $\pm$ 0.01} & {1.23 $\pm$ 0.01} & {1.30 $\pm$ 0.02} & {1.47 $\pm$ 0.04} & {1.91 $\pm$ 0.17} & {3.18 $\pm$ 0.80} \\
{GLRklUCB} & {Worst} & {1.20 $\pm$ 0.01} & {1.34 $\pm$ 0.03} & {1.51 $\pm$ 0.10} & {1.90 $\pm$ 0.35} & {2.84 $\pm$ 0.46} & {5.17 $\pm$ 1.71} \\
{QCD+UCB} & {Unif} & {0.08 $\pm$ 0.01} & {0.07 $\pm$ 0.01} & {0.10 $\pm$ 0.01} & {0.21 $\pm$ 0.04} & {0.59 $\pm$ 0.17} & {1.81 $\pm$ 0.82} \\
{QCD+UCB} & {Worst} & {0.07 $\pm$ 0.00} & {0.09 $\pm$ 0.01} & {0.14 $\pm$ 0.02} & {0.29 $\pm$ 0.06} & {0.59 $\pm$ 0.18} & {1.40 $\pm$ 0.64} \\
{QCD+klUCB} & {Unif} & {1.51 $\pm$ 0.01} & {1.52 $\pm$ 0.01} & {1.56 $\pm$ 0.01} & {1.70 $\pm$ 0.07} & {2.14 $\pm$ 0.19} & {3.46 $\pm$ 0.84} \\
{QCD+klUCB} & {Worst} & {1.50 $\pm$ 0.01} & {1.61 $\pm$ 0.06} & {1.84 $\pm$ 0.24} & {2.37 $\pm$ 0.73} & {3.09 $\pm$ 0.45} & {5.51 $\pm$ 1.80} \\
\hline
\end{tabular}

\vspace{0.5em} 

\begin{tabular}{c|c|cccccc}
\multicolumn{8}{c}{\textbf{(b) Deterministic Change-Points (Columns: $N_C$)}} \\ \hline
\multicolumn{1}{c|}{\bf Algorithm} & \multicolumn{1}{c|}{\bf PB} & \multicolumn{1}{c}{\bf 2938} & \multicolumn{1}{c}{\bf 939} & \multicolumn{1}{c}{\bf 300} & \multicolumn{1}{c}{\bf 96} & \multicolumn{1}{c}{\bf 31} & \multicolumn{1}{c}{\bf 10} \\
\hline
{MASTER} & {Unif} & {2.64 $\pm$ 0.07} & {2.65 $\pm$ 0.08} & {2.64 $\pm$ 0.02} & {2.60 $\pm$ 0.03} & {2.63 $\pm$ 0.09} & {2.64 $\pm$ 0.20} \\
{MASTER} & {Worst} & {2.63 $\pm$ 0.05} & {2.68 $\pm$ 0.09} & {2.65 $\pm$ 0.23} & {2.64 $\pm$ 0.02} & {2.64 $\pm$ 0.16} & {2.67 $\pm$ 0.25} \\
{RR\_p.05} & {Unif} & {1.03 $\pm$ 0.06} & {1.03 $\pm$ 0.04} & {1.01 $\pm$ 0.01} & {1.06 $\pm$ 0.00} & {1.06 $\pm$ 0.08} & {1.00 $\pm$ 0.16} \\
{RR\_p.05} & {Worst} & {0.87 $\pm$ 0.01} & {0.88 $\pm$ 0.05} & {0.89 $\pm$ 0.14} & {0.86 $\pm$ 0.01} & {0.89 $\pm$ 0.06} & {1.12 $\pm$ 0.05} \\
{GLRklUCB} & {Unif} & {1.31 $\pm$ 0.05} & {1.24 $\pm$ 0.05} & {1.27 $\pm$ 0.01} & {1.36 $\pm$ 0.01} & {1.63 $\pm$ 0.08} & {2.38 $\pm$ 0.17} \\
{GLRklUCB} & {Worst} & {1.20 $\pm$ 0.01} & {1.35 $\pm$ 0.07} & {1.60 $\pm$ 0.22} & {1.76 $\pm$ 0.14} & {2.44 $\pm$ 0.27} & {4.36 $\pm$ 1.01} \\
{QCD+UCB} & {Unif} & {0.13 $\pm$ 0.02} & {0.07 $\pm$ 0.02} & {0.08 $\pm$ 0.00} & {0.12 $\pm$ 0.00} & {0.30 $\pm$ 0.04} & {1.03 $\pm$ 0.21} \\
{QCD+UCB} & {Worst} & {0.07 $\pm$ 0.00} & {0.11 $\pm$ 0.03} & {0.13 $\pm$ 0.06} & {0.20 $\pm$ 0.01} & {0.33 $\pm$ 0.01} & {1.22 $\pm$ 0.13} \\
{QCD+klUCB} & {Unif} & {1.55 $\pm$ 0.07} & {1.53 $\pm$ 0.05} & {1.53 $\pm$ 0.01} & {1.59 $\pm$ 0.01} & {1.82 $\pm$ 0.14} & {2.58 $\pm$ 0.14} \\
{QCD+klUCB} & {Worst} & {1.50 $\pm$ 0.00} & {1.61 $\pm$ 0.06} & {2.03 $\pm$ 0.28} & {2.02 $\pm$ 0.18} & {2.78 $\pm$ 0.28} & {4.58 $\pm$ 0.67} \\
\hline
\end{tabular}

\end{center}
\end{table}

\begin{table}[H]
\caption{Average Time Per Run (sec) For Different Change-Points, $T$=100000, 4000 Runs.}
\label{table:time_per_exp_T100000}
\tiny
\setlength{\tabcolsep}{3pt} 
\begin{center}
\begin{tabular}{c|c|cccccc}
\multicolumn{8}{c}{\textbf{(a) Random Change-Points (Columns: $\xi$)}} \\ \hline
\multicolumn{1}{c|}{\bf Algorithm} & \multicolumn{1}{c|}{\bf PB} & \multicolumn{1}{c}{\bf 0.3} & \multicolumn{1}{c}{\bf 0.4} & \multicolumn{1}{c}{\bf 0.5} & \multicolumn{1}{c}{\bf 0.6} & \multicolumn{1}{c}{\bf 0.7} & \multicolumn{1}{c}{\bf 0.8} \\
\hline
{MASTER} & {Unif} & {3.33 $\pm$ 0.03} & {3.34 $\pm$ 0.02} & {3.34 $\pm$ 0.04} & {3.33 $\pm$ 0.03} & {3.36 $\pm$ 0.04} & {3.35 $\pm$ 0.20} \\
{MASTER} & {Worst} & {3.32 $\pm$ 0.03} & {3.32 $\pm$ 0.01} & {3.33 $\pm$ 0.02} & {3.33 $\pm$ 0.05} & {3.35 $\pm$ 0.08} & {3.36 $\pm$ 0.15} \\
{RR\_p.05} & {Unif} & {1.13 $\pm$ 0.02} & {1.12 $\pm$ 0.01} & {1.12 $\pm$ 0.02} & {1.12 $\pm$ 0.03} & {1.12 $\pm$ 0.05} & {1.12 $\pm$ 0.07} \\
{RR\_p.05} & {Worst} & {0.96 $\pm$ 0.01} & {0.94 $\pm$ 0.02} & {0.94 $\pm$ 0.03} & {0.96 $\pm$ 0.05} & {1.03 $\pm$ 0.09} & {1.23 $\pm$ 0.05} \\
{GLRklUCB} & {Unif} & {1.38 $\pm$ 0.01} & {1.37 $\pm$ 0.01} & {1.45 $\pm$ 0.02} & {1.65 $\pm$ 0.05} & {2.18 $\pm$ 0.20} & {3.76 $\pm$ 1.02} \\
{GLRklUCB} & {Worst} & {1.33 $\pm$ 0.03} & {1.49 $\pm$ 0.04} & {1.68 $\pm$ 0.11} & {2.14 $\pm$ 0.37} & {3.27 $\pm$ 0.53} & {6.08 $\pm$ 2.06} \\
{QCD+UCB} & {Unif} & {0.09 $\pm$ 0.01} & {0.08 $\pm$ 0.01} & {0.11 $\pm$ 0.01} & {0.24 $\pm$ 0.04} & {0.72 $\pm$ 0.20} & {2.24 $\pm$ 1.04} \\
{QCD+UCB} & {Worst} & {0.08 $\pm$ 0.01} & {0.10 $\pm$ 0.01} & {0.16 $\pm$ 0.02} & {0.33 $\pm$ 0.07} & {0.68 $\pm$ 0.19} & {1.65 $\pm$ 0.71} \\
{QCD+klUCB} & {Unif} & {1.68 $\pm$ 0.01} & {1.68 $\pm$ 0.01} & {1.73 $\pm$ 0.01} & {1.90 $\pm$ 0.04} & {2.44 $\pm$ 0.22} & {4.12 $\pm$ 1.09} \\
{QCD+klUCB} & {Worst} & {1.67 $\pm$ 0.01} & {1.80 $\pm$ 0.06} & {2.06 $\pm$ 0.28} & {2.71 $\pm$ 0.84} & {3.60 $\pm$ 0.55} & {6.42 $\pm$ 2.13} \\
\hline
\end{tabular}

\vspace{0.5em} 

\begin{tabular}{c|c|cccccc}
\multicolumn{8}{c}{\textbf{(b) Deterministic Change-Points (Columns: $N_C$)}} \\ \hline
\multicolumn{1}{c|}{\bf Algorithm} & \multicolumn{1}{c|}{\bf PB} & \multicolumn{1}{c}{\bf 3163} & \multicolumn{1}{c}{\bf 1000} & \multicolumn{1}{c}{\bf 317} & \multicolumn{1}{c}{\bf 101} & \multicolumn{1}{c}{\bf 32} & \multicolumn{1}{c}{\bf 10} \\
\hline
{MASTER} & {Unif} & {3.35 $\pm$ 0.42} & {3.32 $\pm$ 0.04} & {3.35 $\pm$ 0.22} & {3.32 $\pm$ 0.01} & {3.32 $\pm$ 0.15} & {3.32 $\pm$ 0.02} \\
{MASTER} & {Worst} & {3.38 $\pm$ 0.09} & {3.38 $\pm$ 0.10} & {3.40 $\pm$ 0.08} & {3.38 $\pm$ 0.06} & {3.37 $\pm$ 0.03} & {3.35 $\pm$ 0.07} \\
{RR\_p.05} & {Unif} & {1.15 $\pm$ 0.26} & {1.14 $\pm$ 0.04} & {1.16 $\pm$ 0.12} & {1.12 $\pm$ 0.01} & {1.10 $\pm$ 0.04} & {1.08 $\pm$ 0.11} \\
{RR\_p.05} & {Worst} & {0.97 $\pm$ 0.05} & {0.95 $\pm$ 0.05} & {0.94 $\pm$ 0.02} & {1.00 $\pm$ 0.01} & {1.04 $\pm$ 0.01} & {1.24 $\pm$ 0.01} \\
{GLRklUCB} & {Unif} & {1.46 $\pm$ 0.28} & {1.38 $\pm$ 0.03} & {1.43 $\pm$ 0.17} & {1.52 $\pm$ 0.01} & {1.83 $\pm$ 0.06} & {2.67 $\pm$ 0.20} \\
{GLRklUCB} & {Worst} & {1.33 $\pm$ 0.06} & {1.48 $\pm$ 0.07} & {1.66 $\pm$ 0.06} & {2.04 $\pm$ 0.11} & {2.97 $\pm$ 0.17} & {5.06 $\pm$ 0.94} \\
{QCD+UCB} & {Unif} & {0.12 $\pm$ 0.08} & {0.07 $\pm$ 0.01} & {0.13 $\pm$ 0.05} & {0.13 $\pm$ 0.00} & {0.38 $\pm$ 0.04} & {1.09 $\pm$ 0.14} \\
{QCD+UCB} & {Worst} & {0.09 $\pm$ 0.02} & {0.11 $\pm$ 0.03} & {0.16 $\pm$ 0.02} & {0.41 $\pm$ 0.01} & {0.73 $\pm$ 0.00} & {1.05 $\pm$ 0.03} \\
{QCD+klUCB} & {Unif} & {1.71 $\pm$ 0.29} & {1.69 $\pm$ 0.03} & {1.73 $\pm$ 0.17} & {1.78 $\pm$ 0.01} & {2.02 $\pm$ 0.07} & {2.88 $\pm$ 0.17} \\
{QCD+klUCB} & {Worst} & {1.68 $\pm$ 0.06} & {1.77 $\pm$ 0.07} & {1.95 $\pm$ 0.07} & {2.35 $\pm$ 0.16} & {3.24 $\pm$ 0.22} & {5.09 $\pm$ 0.77} \\
\hline
\end{tabular}

\end{center}
\end{table}

\begin{table}[H]
\caption{Mean Number Of Declared Changes For Different Change-Points, $T=1000$, 4000 Runs.}
\label{table:change_detect_T1000}
\scriptsize
\setlength{\tabcolsep}{3pt} 
\begin{center}
\begin{tabular}{c|c|cccccc}
\multicolumn{8}{c}{\textbf{(a) Random Change-Points (Columns: $\xi$)}} \\ \hline
\multicolumn{1}{c|}{\bf Algorithm} & \multicolumn{1}{c|}{\bf PB} & \multicolumn{1}{c}{\bf 0.3} & \multicolumn{1}{c}{\bf 0.4} & \multicolumn{1}{c}{\bf 0.5} & \multicolumn{1}{c}{\bf 0.6} & \multicolumn{1}{c}{\bf 0.7} & \multicolumn{1}{c}{\bf 0.8} \\
\hline
{MASTER} & {Unif} & 0.00 & 0.00 & 0.00 & 0.00 & 0.00 & 0.00 \\
{MASTER} & {Worst} & 0.00 & 0.00 & 0.00 & 0.00 & 0.00 & 0.00 \\
{GLRklUCB} & {Unif} & 0.54 & 1.09 & 1.39 & 1.39 & 1.19 & 0.89 \\
{GLRklUCB} & {Worst} & 2.24 & 1.80 & 1.12 & 0.24 & 0.02 & 0.00 \\
{QCD+UCB} & {Unif} & 0.45 & 1.02 & 1.40 & 1.47 & 1.26 & 0.98 \\
{QCD+UCB} & {Worst} & 2.25 & 1.93 & 1.18 & 0.37 & 0.03 & 0.01 \\
{QCD+klUCB} & {Unif} & 0.70 & 1.20 & 1.42 & 1.36 & 1.11 & 0.82 \\
{QCD+klUCB} & {Worst} & 2.59 & 1.86 & 1.13 & 0.23 & 0.01 & 0.00 \\
\hline
\end{tabular}

\vspace{0.5em} 

\begin{tabular}{c|c|cccccc}
\multicolumn{8}{c}{\textbf{(b) Deterministic Change-Points (Columns: $N_C$)}} \\ \hline
\multicolumn{1}{c|}{\bf Algorithm} & \multicolumn{1}{c|}{\bf PB} & \multicolumn{1}{c}{\bf 126} & \multicolumn{1}{c}{\bf 64} & \multicolumn{1}{c}{\bf 32} & \multicolumn{1}{c}{\bf 16} & \multicolumn{1}{c}{\bf 8} & \multicolumn{1}{c}{\bf 4} \\
\hline
{MASTER} & {Unif} & 0.00 & 0.00 & 0.00 & 0.00 & 0.00 & 0.00 \\
{MASTER} & {Worst} & 0.00 & 0.00 & 0.00 & 0.00 & 0.00 & 0.00 \\
{GLRklUCB} & {Unif} & 0.25 & 0.57 & 0.52 & 0.78 & 1.07 & 1.15 \\
{GLRklUCB} & {Worst} & 2.56 & 2.06 & 1.00 & 0.51 & 0.00 & 0.00 \\
{QCD+UCB} & {Unif} & 0.00 & 1.00 & 0.00 & 0.00 & 1.00 & 1.00 \\
{QCD+UCB} & {Worst} & 3.00 & 2.00 & 1.00 & 0.00 & 0.00 & 0.00 \\
{QCD+klUCB} & {Unif} & 0.41 & 0.55 & 0.94 & 0.72 & 1.05 & 1.16 \\
{QCD+klUCB} & {Worst} & 2.88 & 2.04 & 1.01 & 0.26 & 0.00 & 0.00 \\
\hline
\end{tabular}

\end{center}
\end{table}

\begin{table}[H]
\caption{Mean Number Of Declared Changes For Different Change-Points, $T=2000$, 4000 Runs.}
\label{table:change_detect_T2000}
\scriptsize
\setlength{\tabcolsep}{3pt} 
\begin{center}
\begin{tabular}{c|c|cccccc}
\multicolumn{8}{c}{\textbf{(a) Random Change-Points (Columns: $\xi$)}} \\ \hline
\multicolumn{1}{c|}{\bf Algorithm} & \multicolumn{1}{c|}{\bf PB} & \multicolumn{1}{c}{\bf 0.3} & \multicolumn{1}{c}{\bf 0.4} & \multicolumn{1}{c}{\bf 0.5} & \multicolumn{1}{c}{\bf 0.6} & \multicolumn{1}{c}{\bf 0.7} & \multicolumn{1}{c}{\bf 0.8} \\
\hline
{MASTER} & {Unif} & 0.00 & 0.00 & 0.00 & 0.00 & 0.00 & 0.00 \\
{MASTER} & {Worst} & 0.00 & 0.00 & 0.00 & 0.00 & 0.00 & 0.00 \\
{GLRklUCB} & {Unif} & 1.44 & 2.65 & 3.17 & 2.95 & 2.31 & 1.60 \\
{GLRklUCB} & {Worst} & 4.71 & 3.59 & 2.07 & 0.90 & 0.14 & 0.01 \\
{QCD+UCB} & {Unif} & 1.16 & 2.46 & 3.14 & 3.04 & 2.44 & 1.70 \\
{QCD+UCB} & {Worst} & 4.59 & 3.67 & 2.21 & 1.01 & 0.25 & 0.03 \\
{QCD+klUCB} & {Unif} & 1.71 & 2.79 & 3.05 & 2.73 & 2.11 & 1.42 \\
{QCD+klUCB} & {Worst} & 5.20 & 3.36 & 1.84 & 0.88 & 0.15 & 0.01 \\
\hline
\end{tabular}

\vspace{0.5em} 

\begin{tabular}{c|c|cccccc}
\multicolumn{8}{c}{\textbf{(b) Deterministic Change-Points (Columns: $N_C$)}} \\ \hline
\multicolumn{1}{c|}{\bf Algorithm} & \multicolumn{1}{c|}{\bf PB} & \multicolumn{1}{c}{\bf 205} & \multicolumn{1}{c}{\bf 96} & \multicolumn{1}{c}{\bf 45} & \multicolumn{1}{c}{\bf 21} & \multicolumn{1}{c}{\bf 10} & \multicolumn{1}{c}{\bf 5} \\
\hline
{MASTER} & {Unif} & 0.00 & 0.00 & 0.00 & 0.00 & 0.00 & 0.00 \\
{MASTER} & {Worst} & 0.00 & 0.00 & 0.00 & 0.00 & 0.00 & 0.00 \\
{GLRklUCB} & {Unif} & 1.03 & 0.96 & 2.65 & 3.24 & 2.52 & 1.98 \\
{GLRklUCB} & {Worst} & 5.13 & 3.15 & 2.30 & 1.01 & 0.14 & 0.00 \\
{QCD+UCB} & {Unif} & 1.00 & 1.00 & 3.00 & 3.00 & 4.00 & 2.00 \\
{QCD+UCB} & {Worst} & 5.00 & 3.00 & 3.00 & 1.00 & 0.00 & 0.00 \\
{QCD+klUCB} & {Unif} & 0.85 & 1.01 & 2.74 & 3.05 & 3.19 & 2.23 \\
{QCD+klUCB} & {Worst} & 5.53 & 3.01 & 2.11 & 1.00 & 0.17 & 0.00 \\
\hline
\end{tabular}

\end{center}
\end{table}

\begin{table}[H]
\caption{Mean Number Of Declared Changes For Different Change-Points, $T=5000$, 4000 Runs.}
\label{table:change_detect_T5000}
\scriptsize
\setlength{\tabcolsep}{3pt} 
\begin{center}
\begin{tabular}{c|c|cccccc}
\multicolumn{8}{c}{\textbf{(a) Random Change-Points (Columns: $\xi$)}} \\ \hline
\multicolumn{1}{c|}{\bf Algorithm} & \multicolumn{1}{c|}{\bf PB} & \multicolumn{1}{c}{\bf 0.3} & \multicolumn{1}{c}{\bf 0.4} & \multicolumn{1}{c}{\bf 0.5} & \multicolumn{1}{c}{\bf 0.6} & \multicolumn{1}{c}{\bf 0.7} & \multicolumn{1}{c}{\bf 0.8} \\
\hline
{MASTER} & {Unif} & 0.00 & 0.00 & 0.00 & 0.00 & 0.00 & 0.00 \\
{MASTER} & {Worst} & 0.00 & 0.00 & 0.00 & 0.00 & 0.00 & 0.00 \\
{GLRklUCB} & {Unif} & 4.69 & 7.46 & 7.80 & 6.44 & 4.55 & 2.77 \\
{GLRklUCB} & {Worst} & 11.37 & 7.56 & 4.42 & 2.26 & 0.72 & 0.11 \\
{QCD+UCB} & {Unif} & 3.89 & 7.17 & 7.98 & 6.80 & 4.89 & 3.02 \\
{QCD+UCB} & {Worst} & 11.08 & 7.91 & 5.51 & 2.82 & 0.88 & 0.26 \\
{QCD+klUCB} & {Unif} & 5.25 & 7.59 & 7.44 & 6.03 & 4.13 & 2.50 \\
{QCD+klUCB} & {Worst} & 11.88 & 6.72 & 3.71 & 2.07 & 0.71 & 0.09 \\
\hline
\end{tabular}

\vspace{0.5em} 

\begin{tabular}{c|c|cccccc}
\multicolumn{8}{c}{\textbf{(b) Deterministic Change-Points (Columns: $N_C$)}} \\ \hline
\multicolumn{1}{c|}{\bf Algorithm} & \multicolumn{1}{c|}{\bf PB} & \multicolumn{1}{c}{\bf 389} & \multicolumn{1}{c}{\bf 166} & \multicolumn{1}{c}{\bf 71} & \multicolumn{1}{c}{\bf 31} & \multicolumn{1}{c}{\bf 13} & \multicolumn{1}{c}{\bf 6} \\
\hline
{MASTER} & {Unif} & 0.00 & 0.00 & 0.00 & 0.00 & 0.00 & 0.00 \\
{MASTER} & {Worst} & 0.00 & 0.00 & 0.00 & 0.00 & 0.00 & 0.00 \\
{GLRklUCB} & {Unif} & 3.36 & 5.19 & 9.05 & 8.86 & 5.38 & 3.85 \\
{GLRklUCB} & {Worst} & 11.54 & 8.56 & 5.05 & 3.03 & 0.98 & 0.02 \\
{QCD+UCB} & {Unif} & 1.00 & 3.00 & 9.00 & 9.00 & 7.00 & 3.00 \\
{QCD+UCB} & {Worst} & 9.00 & 8.00 & 5.00 & 4.00 & 1.00 & 1.00 \\
{QCD+klUCB} & {Unif} & 3.99 & 5.20 & 8.98 & 8.37 & 4.08 & 3.12 \\
{QCD+klUCB} & {Worst} & 12.98 & 8.57 & 5.18 & 3.01 & 0.98 & 0.06 \\
\hline
\end{tabular}

\end{center}
\end{table}

\begin{table}[H]
\caption{Mean Number Of Declared Changes For Different Change-Points, $T=10000$, 4000 Runs.}
\label{table:change_detect_T10000}
\scriptsize
\setlength{\tabcolsep}{3pt} 
\begin{center}
\begin{tabular}{c|c|cccccc}
\multicolumn{8}{c}{\textbf{(a) Random Change-Points (Columns: $\xi$)}} \\ \hline
\multicolumn{1}{c|}{\bf Algorithm} & \multicolumn{1}{c|}{\bf PB} & \multicolumn{1}{c}{\bf 0.3} & \multicolumn{1}{c}{\bf 0.4} & \multicolumn{1}{c}{\bf 0.5} & \multicolumn{1}{c}{\bf 0.6} & \multicolumn{1}{c}{\bf 0.7} & \multicolumn{1}{c}{\bf 0.8} \\
\hline
{MASTER} & {Unif} & 0.00 & 0.00 & 0.00 & 0.00 & 0.00 & 0.00 \\
{MASTER} & {Worst} & 0.00 & 0.00 & 0.00 & 0.00 & 0.00 & 0.00 \\
{GLRklUCB} & {Unif} & 10.57 & 15.23 & 14.61 & 11.30 & 7.18 & 3.94 \\
{GLRklUCB} & {Worst} & 21.21 & 12.99 & 8.01 & 4.49 & 1.20 & 0.31 \\
{QCD+UCB} & {Unif} & 9.02 & 14.99 & 15.10 & 11.88 & 7.66 & 4.22 \\
{QCD+UCB} & {Worst} & 20.93 & 14.03 & 11.00 & 6.24 & 1.52 & 0.64 \\
{QCD+klUCB} & {Unif} & 11.60 & 15.25 & 13.91 & 10.36 & 6.43 & 3.50 \\
{QCD+klUCB} & {Worst} & 21.30 & 11.12 & 6.34 & 3.41 & 1.17 & 0.28 \\
\hline
\end{tabular}

\vspace{0.5em} 

\begin{tabular}{c|c|cccccc}
\multicolumn{8}{c}{\textbf{(b) Deterministic Change-Points (Columns: $N_C$)}} \\ \hline
\multicolumn{1}{c|}{\bf Algorithm} & \multicolumn{1}{c|}{\bf PB} & \multicolumn{1}{c}{\bf 631} & \multicolumn{1}{c}{\bf 252} & \multicolumn{1}{c}{\bf 100} & \multicolumn{1}{c}{\bf 40} & \multicolumn{1}{c}{\bf 16} & \multicolumn{1}{c}{\bf 7} \\
\hline
{MASTER} & {Unif} & 0.00 & 0.00 & 0.00 & 0.00 & 0.00 & 0.00 \\
{MASTER} & {Worst} & 0.00 & 0.00 & 0.00 & 0.00 & 0.00 & 0.00 \\
{GLRklUCB} & {Unif} & 4.84 & 11.79 & 14.11 & 14.58 & 7.65 & 5.44 \\
{GLRklUCB} & {Worst} & 21.31 & 11.93 & 6.31 & 5.66 & 1.23 & 0.18 \\
{QCD+UCB} & {Unif} & 4.00 & 13.00 & 13.00 & 16.00 & 11.00 & 6.00 \\
{QCD+UCB} & {Worst} & 22.00 & 12.00 & 8.00 & 12.00 & 2.00 & 1.00 \\
{QCD+klUCB} & {Unif} & 5.25 & 11.81 & 13.96 & 13.70 & 7.03 & 4.75 \\
{QCD+klUCB} & {Worst} & 22.23 & 9.64 & 5.15 & 3.79 & 1.24 & 0.11 \\
\hline
\end{tabular}

\end{center}
\end{table}

\begin{table}[H]
\caption{Mean Number Of Declared Changes For Different Change-Points, $T=20000$, 4000 Runs.}
\label{table:change_detect_T20000}
\scriptsize
\setlength{\tabcolsep}{3pt} 
\begin{center}
\begin{tabular}{c|c|cccccc}
\multicolumn{8}{c}{\textbf{(a) Random Change-Points (Columns: $\xi$)}} \\ \hline
\multicolumn{1}{c|}{\bf Algorithm} & \multicolumn{1}{c|}{\bf PB} & \multicolumn{1}{c}{\bf 0.3} & \multicolumn{1}{c}{\bf 0.4} & \multicolumn{1}{c}{\bf 0.5} & \multicolumn{1}{c}{\bf 0.6} & \multicolumn{1}{c}{\bf 0.7} & \multicolumn{1}{c}{\bf 0.8} \\
\hline
{MASTER} & {Unif} & 0.00 & 0.00 & 0.00 & 0.00 & 0.00 & 0.00 \\
{MASTER} & {Worst} & 0.00 & 0.00 & 0.00 & 0.00 & 0.00 & 0.00 \\
{GLRklUCB} & {Unif} & 23.18 & 30.38 & 26.91 & 18.80 & 10.69 & 5.21 \\
{GLRklUCB} & {Worst} & 38.72 & 21.87 & 13.59 & 7.26 & 1.92 & 0.56 \\
{QCD+UCB} & {Unif} & 20.43 & 30.33 & 27.88 & 20.02 & 11.56 & 5.55 \\
{QCD+UCB} & {Worst} & 38.69 & 25.34 & 20.58 & 11.81 & 2.78 & 1.17 \\
{QCD+klUCB} & {Unif} & 25.08 & 29.86 & 25.19 & 17.22 & 9.63 & 4.68 \\
{QCD+klUCB} & {Worst} & 37.43 & 18.71 & 10.61 & 5.20 & 1.85 & 0.51 \\
\hline
\end{tabular}

\vspace{0.5em} 

\begin{tabular}{c|c|cccccc}
\multicolumn{8}{c}{\textbf{(b) Deterministic Change-Points (Columns: $N_C$)}} \\ \hline
\multicolumn{1}{c|}{\bf Algorithm} & \multicolumn{1}{c|}{\bf PB} & \multicolumn{1}{c}{\bf 1025} & \multicolumn{1}{c}{\bf 381} & \multicolumn{1}{c}{\bf 142} & \multicolumn{1}{c}{\bf 53} & \multicolumn{1}{c}{\bf 20} & \multicolumn{1}{c}{\bf 8} \\
\hline
{MASTER} & {Unif} & 0.00 & 0.00 & 0.00 & 0.00 & 0.00 & 0.00 \\
{MASTER} & {Worst} & 0.00 & 0.00 & 0.00 & 0.00 & 0.00 & 0.00 \\
{GLRklUCB} & {Unif} & 14.18 & 28.21 & 28.32 & 21.07 & 12.20 & 8.24 \\
{GLRklUCB} & {Worst} & 35.36 & 20.82 & 15.04 & 12.29 & 2.11 & 1.02 \\
{QCD+UCB} & {Unif} & 11.00 & 27.00 & 29.00 & 21.00 & 15.00 & 9.00 \\
{QCD+UCB} & {Worst} & 37.00 & 26.00 & 26.00 & 25.00 & 3.00 & 1.00 \\
{QCD+klUCB} & {Unif} & 14.94 & 27.12 & 26.18 & 19.49 & 10.57 & 8.32 \\
{QCD+klUCB} & {Worst} & 33.97 & 16.30 & 11.05 & 6.62 & 2.08 & 0.93 \\
\hline
\end{tabular}

\end{center}
\end{table}

\begin{table}[H]
\caption{Mean Number Of Declared Changes For Different Change-Points, $T=30000$, 4000 Runs.}
\label{table:change_detect_T30000}
\scriptsize
\setlength{\tabcolsep}{3pt} 
\begin{center}
\begin{tabular}{c|c|cccccc}
\multicolumn{8}{c}{\textbf{(a) Random Change-Points (Columns: $\xi$)}} \\ \hline
\multicolumn{1}{c|}{\bf Algorithm} & \multicolumn{1}{c|}{\bf PB} & \multicolumn{1}{c}{\bf 0.3} & \multicolumn{1}{c}{\bf 0.4} & \multicolumn{1}{c}{\bf 0.5} & \multicolumn{1}{c}{\bf 0.6} & \multicolumn{1}{c}{\bf 0.7} & \multicolumn{1}{c}{\bf 0.8} \\
\hline
{MASTER} & {Unif} & 0.00 & 0.00 & 0.00 & 0.00 & 0.00 & 0.00 \\
{MASTER} & {Worst} & 0.00 & 0.00 & 0.00 & 0.00 & 0.00 & 0.00 \\
{GLRklUCB} & {Unif} & 36.25 & 44.93 & 37.85 & 24.82 & 13.41 & 6.11 \\
{GLRklUCB} & {Worst} & 54.32 & 30.20 & 18.53 & 9.39 & 2.54 & 0.75 \\
{QCD+UCB} & {Unif} & 32.46 & 45.20 & 39.44 & 26.50 & 14.45 & 6.52 \\
{QCD+UCB} & {Worst} & 54.72 & 36.38 & 29.41 & 16.02 & 4.02 & 1.62 \\
{QCD+klUCB} & {Unif} & 38.81 & 43.89 & 35.38 & 22.59 & 12.04 & 5.54 \\
{QCD+klUCB} & {Worst} & 51.41 & 25.41 & 14.02 & 6.68 & 2.36 & 0.69 \\
\hline
\end{tabular}

\vspace{0.5em} 

\begin{tabular}{c|c|cccccc}
\multicolumn{8}{c}{\textbf{(b) Deterministic Change-Points (Columns: $N_C$)}} \\ \hline
\multicolumn{1}{c|}{\bf Algorithm} & \multicolumn{1}{c|}{\bf PB} & \multicolumn{1}{c}{\bf 1362} & \multicolumn{1}{c}{\bf 486} & \multicolumn{1}{c}{\bf 174} & \multicolumn{1}{c}{\bf 62} & \multicolumn{1}{c}{\bf 23} & \multicolumn{1}{c}{\bf 8} \\
\hline
{MASTER} & {Unif} & 0.00 & 0.00 & 0.00 & 0.00 & 0.00 & 0.00 \\
{MASTER} & {Worst} & 0.00 & 0.00 & 0.00 & 0.00 & 0.00 & 0.00 \\
{GLRklUCB} & {Unif} & 25.80 & 42.86 & 38.75 & 29.42 & 16.60 & 6.73 \\
{GLRklUCB} & {Worst} & 50.45 & 29.63 & 17.89 & 8.95 & 2.05 & 1.01 \\
{QCD+UCB} & {Unif} & 21.00 & 40.00 & 42.00 & 32.00 & 18.00 & 8.00 \\
{QCD+UCB} & {Worst} & 47.00 & 34.00 & 20.00 & 13.00 & 3.00 & 1.00 \\
{QCD+klUCB} & {Unif} & 28.17 & 41.30 & 36.74 & 27.47 & 14.55 & 5.13 \\
{QCD+klUCB} & {Worst} & 50.05 & 24.27 & 16.81 & 7.82 & 2.03 & 1.01 \\
\hline
\end{tabular}

\end{center}
\end{table}

\begin{table}[H]
\caption{Mean Number Of Declared Changes For Different Change-Points, $T=40000$, 4000 Runs.}
\label{table:change_detect_T40000}
\scriptsize
\setlength{\tabcolsep}{3pt} 
\begin{center}
\begin{tabular}{c|c|cccccc}
\multicolumn{8}{c}{\textbf{(a) Random Change-Points (Columns: $\xi$)}} \\ \hline
\multicolumn{1}{c|}{\bf Algorithm} & \multicolumn{1}{c|}{\bf PB} & \multicolumn{1}{c}{\bf 0.3} & \multicolumn{1}{c}{\bf 0.4} & \multicolumn{1}{c}{\bf 0.5} & \multicolumn{1}{c}{\bf 0.6} & \multicolumn{1}{c}{\bf 0.7} & \multicolumn{1}{c}{\bf 0.8} \\
\hline
{MASTER} & {Unif} & 0.00 & 0.00 & 0.00 & 0.00 & 0.00 & 0.00 \\
{MASTER} & {Worst} & 0.00 & 0.00 & 0.00 & 0.00 & 0.00 & 0.00 \\
{GLRklUCB} & {Unif} & 49.12 & 58.96 & 47.95 & 30.22 & 15.50 & 6.74 \\
{GLRklUCB} & {Worst} & 69.09 & 37.81 & 22.92 & 11.65 & 3.14 & 0.88 \\
{QCD+UCB} & {Unif} & 44.89 & 59.45 & 49.92 & 32.30 & 16.69 & 7.13 \\
{QCD+UCB} & {Worst} & 69.89 & 47.26 & 37.66 & 19.92 & 5.29 & 2.02 \\
{QCD+klUCB} & {Unif} & 52.45 & 57.35 & 44.60 & 27.43 & 13.98 & 6.14 \\
{QCD+klUCB} & {Worst} & 64.55 & 31.68 & 17.12 & 7.91 & 2.81 & 0.80 \\
\hline
\end{tabular}

\vspace{0.5em} 

\begin{tabular}{c|c|cccccc}
\multicolumn{8}{c}{\textbf{(b) Deterministic Change-Points (Columns: $N_C$)}} \\ \hline
\multicolumn{1}{c|}{\bf Algorithm} & \multicolumn{1}{c|}{\bf PB} & \multicolumn{1}{c}{\bf 1666} & \multicolumn{1}{c}{\bf 578} & \multicolumn{1}{c}{\bf 200} & \multicolumn{1}{c}{\bf 70} & \multicolumn{1}{c}{\bf 25} & \multicolumn{1}{c}{\bf 9} \\
\hline
{MASTER} & {Unif} & 0.00 & 0.00 & 0.00 & 0.00 & 0.00 & 0.00 \\
{MASTER} & {Worst} & 0.00 & 0.00 & 0.00 & 0.00 & 0.00 & 0.00 \\
{GLRklUCB} & {Unif} & 35.21 & 56.62 & 51.68 & 37.09 & 20.23 & 7.43 \\
{GLRklUCB} & {Worst} & 66.30 & 33.00 & 26.14 & 11.18 & 5.94 & 0.83 \\
{QCD+UCB} & {Unif} & 32.00 & 60.00 & 51.00 & 39.00 & 21.00 & 8.00 \\
{QCD+UCB} & {Worst} & 68.00 & 40.00 & 46.00 & 21.00 & 11.00 & 3.00 \\
{QCD+klUCB} & {Unif} & 38.36 & 56.47 & 49.78 & 35.09 & 18.82 & 6.55 \\
{QCD+klUCB} & {Worst} & 64.75 & 28.52 & 19.85 & 8.88 & 3.36 & 0.79 \\
\hline
\end{tabular}

\end{center}
\end{table}

\begin{table}[H]
\caption{Mean Number Of Declared Changes For Different Change-Points, $T=50000$, 4000 Runs.}
\label{table:change_detect_T50000}
\scriptsize
\setlength{\tabcolsep}{3pt} 
\begin{center}
\begin{tabular}{c|c|cccccc}
\multicolumn{8}{c}{\textbf{(a) Random Change-Points (Columns: $\xi$)}} \\ \hline
\multicolumn{1}{c|}{\bf Algorithm} & \multicolumn{1}{c|}{\bf PB} & \multicolumn{1}{c}{\bf 0.3} & \multicolumn{1}{c}{\bf 0.4} & \multicolumn{1}{c}{\bf 0.5} & \multicolumn{1}{c}{\bf 0.6} & \multicolumn{1}{c}{\bf 0.7} & \multicolumn{1}{c}{\bf 0.8} \\
\hline
{MASTER} & {Unif} & 0.00 & 0.00 & 0.00 & 0.00 & 0.00 & 0.00 \\
{MASTER} & {Worst} & 0.00 & 0.00 & 0.00 & 0.00 & 0.00 & 0.00 \\
{GLRklUCB} & {Unif} & 62.57 & 72.53 & 57.36 & 34.97 & 17.32 & 7.30 \\
{GLRklUCB} & {Worst} & 83.18 & 45.08 & 27.44 & 13.90 & 3.71 & 1.01 \\
{QCD+UCB} & {Unif} & 57.48 & 73.43 & 59.88 & 37.49 & 18.54 & 7.64 \\
{QCD+UCB} & {Worst} & 84.42 & 58.05 & 45.35 & 23.09 & 6.44 & 2.36 \\
{QCD+klUCB} & {Unif} & 66.14 & 70.39 & 53.19 & 31.69 & 15.57 & 6.62 \\
{QCD+klUCB} & {Worst} & 76.85 & 37.52 & 19.82 & 8.98 & 3.18 & 0.92 \\
\hline
\end{tabular}

\vspace{0.5em} 

\begin{tabular}{c|c|cccccc}
\multicolumn{8}{c}{\textbf{(b) Deterministic Change-Points (Columns: $N_C$)}} \\ \hline
\multicolumn{1}{c|}{\bf Algorithm} & \multicolumn{1}{c|}{\bf PB} & \multicolumn{1}{c}{\bf 1947} & \multicolumn{1}{c}{\bf 660} & \multicolumn{1}{c}{\bf 224} & \multicolumn{1}{c}{\bf 76} & \multicolumn{1}{c}{\bf 26} & \multicolumn{1}{c}{\bf 9} \\
\hline
{MASTER} & {Unif} & 0.00 & 0.00 & 0.00 & 0.00 & 0.00 & 0.00 \\
{MASTER} & {Worst} & 0.00 & 0.00 & 0.00 & 0.00 & 0.00 & 0.00 \\
{GLRklUCB} & {Unif} & 43.44 & 73.24 & 67.07 & 48.33 & 26.89 & 9.00 \\
{GLRklUCB} & {Worst} & 86.39 & 43.53 & 26.07 & 19.61 & 6.61 & 1.11 \\
{QCD+UCB} & {Unif} & 40.00 & 73.00 & 72.00 & 50.00 & 30.00 & 9.00 \\
{QCD+UCB} & {Worst} & 87.00 & 52.00 & 56.00 & 48.00 & 12.00 & 3.00 \\
{QCD+klUCB} & {Unif} & 45.52 & 71.44 & 65.64 & 42.99 & 23.87 & 8.94 \\
{QCD+klUCB} & {Worst} & 82.58 & 39.78 & 17.64 & 8.70 & 4.13 & 1.08 \\
\hline
\end{tabular}

\end{center}
\end{table}

\begin{table}[H]
\caption{Mean Number Of Declared Changes For Different Change-Points, $T=60000$, 4000 Runs.}
\label{table:change_detect_T60000}
\scriptsize
\setlength{\tabcolsep}{3pt} 
\begin{center}
\begin{tabular}{c|c|cccccc}
\multicolumn{8}{c}{\textbf{(a) Random Change-Points (Columns: $\xi$)}} \\ \hline
\multicolumn{1}{c|}{\bf Algorithm} & \multicolumn{1}{c|}{\bf PB} & \multicolumn{1}{c}{\bf 0.3} & \multicolumn{1}{c}{\bf 0.4} & \multicolumn{1}{c}{\bf 0.5} & \multicolumn{1}{c}{\bf 0.6} & \multicolumn{1}{c}{\bf 0.7} & \multicolumn{1}{c}{\bf 0.8} \\
\hline
{MASTER} & {Unif} & 0.00 & 0.00 & 0.00 & 0.00 & 0.00 & 0.00 \\
{MASTER} & {Worst} & 0.00 & 0.00 & 0.00 & 0.00 & 0.00 & 0.00 \\
{GLRklUCB} & {Unif} & 75.81 & 86.09 & 66.55 & 39.34 & 18.77 & 7.68 \\
{GLRklUCB} & {Worst} & 97.03 & 52.35 & 31.42 & 15.38 & 4.25 & 1.12 \\
{QCD+UCB} & {Unif} & 70.20 & 86.99 & 69.40 & 42.17 & 20.12 & 8.02 \\
{QCD+UCB} & {Worst} & 98.60 & 68.75 & 52.67 & 26.29 & 7.63 & 2.71 \\
{QCD+klUCB} & {Unif} & 79.83 & 83.07 & 61.47 & 35.55 & 16.96 & 7.03 \\
{QCD+klUCB} & {Worst} & 88.81 & 43.35 & 22.48 & 9.95 & 3.52 & 1.02 \\
\hline
\end{tabular}

\vspace{0.5em} 

\begin{tabular}{c|c|cccccc}
\multicolumn{8}{c}{\textbf{(b) Deterministic Change-Points (Columns: $N_C$)}} \\ \hline
\multicolumn{1}{c|}{\bf Algorithm} & \multicolumn{1}{c|}{\bf PB} & \multicolumn{1}{c}{\bf 2212} & \multicolumn{1}{c}{\bf 737} & \multicolumn{1}{c}{\bf 245} & \multicolumn{1}{c}{\bf 82} & \multicolumn{1}{c}{\bf 28} & \multicolumn{1}{c}{\bf 10} \\
\hline
{MASTER} & {Unif} & 0.00 & 0.00 & 0.00 & 0.00 & 0.00 & 0.00 \\
{MASTER} & {Worst} & 0.00 & 0.00 & 0.00 & 0.00 & 0.00 & 0.00 \\
{GLRklUCB} & {Unif} & 53.48 & 79.56 & 71.50 & 49.25 & 25.33 & 8.84 \\
{GLRklUCB} & {Worst} & 95.25 & 52.85 & 34.92 & 11.68 & 7.03 & 1.14 \\
{QCD+UCB} & {Unif} & 46.00 & 86.00 & 73.00 & 54.00 & 28.00 & 9.00 \\
{QCD+UCB} & {Worst} & 95.00 & 62.00 & 67.00 & 18.00 & 13.00 & 3.00 \\
{QCD+klUCB} & {Unif} & 58.17 & 79.64 & 66.73 & 45.14 & 22.18 & 8.01 \\
{QCD+klUCB} & {Worst} & 85.25 & 46.06 & 22.47 & 10.82 & 4.79 & 1.08 \\
\hline
\end{tabular}

\end{center}
\end{table}

\begin{table}[H]
\caption{Mean Number Of Declared Changes For Different Change-Points, $T=70000$, 4000 Runs.}
\label{table:change_detect_T70000}
\scriptsize
\setlength{\tabcolsep}{3pt} 
\begin{center}
\begin{tabular}{c|c|cccccc}
\multicolumn{8}{c}{\textbf{(a) Random Change-Points (Columns: $\xi$)}} \\ \hline
\multicolumn{1}{c|}{\bf Algorithm} & \multicolumn{1}{c|}{\bf PB} & \multicolumn{1}{c}{\bf 0.3} & \multicolumn{1}{c}{\bf 0.4} & \multicolumn{1}{c}{\bf 0.5} & \multicolumn{1}{c}{\bf 0.6} & \multicolumn{1}{c}{\bf 0.7} & \multicolumn{1}{c}{\bf 0.8} \\
\hline
{MASTER} & {Unif} & 0.00 & 0.00 & 0.00 & 0.00 & 0.00 & 0.00 \\
{MASTER} & {Worst} & 0.00 & 0.00 & 0.00 & 0.00 & 0.00 & 0.00 \\
{GLRklUCB} & {Unif} & 88.99 & 99.08 & 75.13 & 43.34 & 20.11 & 8.02 \\
{GLRklUCB} & {Worst} & 109.87 & 58.83 & 34.78 & 16.57 & 4.71 & 1.20 \\
{QCD+UCB} & {Unif} & 82.91 & 100.62 & 78.58 & 46.57 & 21.56 & 8.37 \\
{QCD+UCB} & {Worst} & 112.11 & 79.35 & 59.54 & 29.00 & 8.70 & 3.01 \\
{QCD+klUCB} & {Unif} & 93.44 & 95.51 & 69.35 & 39.19 & 18.23 & 7.36 \\
{QCD+klUCB} & {Worst} & 100.06 & 48.65 & 24.90 & 10.78 & 3.82 & 1.07 \\
\hline
\end{tabular}

\vspace{0.5em} 

\begin{tabular}{c|c|cccccc}
\multicolumn{8}{c}{\textbf{(b) Deterministic Change-Points (Columns: $N_C$)}} \\ \hline
\multicolumn{1}{c|}{\bf Algorithm} & \multicolumn{1}{c|}{\bf PB} & \multicolumn{1}{c}{\bf 2464} & \multicolumn{1}{c}{\bf 808} & \multicolumn{1}{c}{\bf 265} & \multicolumn{1}{c}{\bf 87} & \multicolumn{1}{c}{\bf 29} & \multicolumn{1}{c}{\bf 10} \\
\hline
{MASTER} & {Unif} & 0.00 & 0.00 & 0.00 & 0.00 & 0.00 & 0.00 \\
{MASTER} & {Worst} & 0.00 & 0.00 & 0.00 & 0.00 & 0.00 & 0.00 \\
{GLRklUCB} & {Unif} & 63.38 & 97.11 & 90.14 & 53.90 & 25.12 & 10.82 \\
{GLRklUCB} & {Worst} & 105.55 & 51.41 & 38.35 & 18.39 & 3.67 & 1.55 \\
{QCD+UCB} & {Unif} & 47.00 & 95.00 & 96.00 & 57.00 & 27.00 & 11.00 \\
{QCD+UCB} & {Worst} & 100.00 & 75.00 & 66.00 & 33.00 & 7.00 & 4.00 \\
{QCD+klUCB} & {Unif} & 68.35 & 93.91 & 85.27 & 48.62 & 23.75 & 10.01 \\
{QCD+klUCB} & {Worst} & 97.04 & 40.04 & 25.01 & 12.63 & 3.52 & 1.47 \\
\hline
\end{tabular}

\end{center}
\end{table}

\begin{table}[H]
\caption{Mean Number Of Declared Changes For Different Change-Points, $T=80000$, 4000 Runs.}
\label{table:change_detect_T80000}
\scriptsize
\setlength{\tabcolsep}{3pt} 
\begin{center}
\begin{tabular}{c|c|cccccc}
\multicolumn{8}{c}{\textbf{(a) Random Change-Points (Columns: $\xi$)}} \\ \hline
\multicolumn{1}{c|}{\bf Algorithm} & \multicolumn{1}{c|}{\bf PB} & \multicolumn{1}{c}{\bf 0.3} & \multicolumn{1}{c}{\bf 0.4} & \multicolumn{1}{c}{\bf 0.5} & \multicolumn{1}{c}{\bf 0.6} & \multicolumn{1}{c}{\bf 0.7} & \multicolumn{1}{c}{\bf 0.8} \\
\hline
{MASTER} & {Unif} & 0.00 & 0.00 & 0.00 & 0.00 & 0.00 & 0.00 \\
{MASTER} & {Worst} & 0.00 & 0.00 & 0.00 & 0.00 & 0.00 & 0.00 \\
{GLRklUCB} & {Unif} & 102.41 & 112.32 & 83.45 & 47.05 & 21.56 & 8.32 \\
{GLRklUCB} & {Worst} & 122.49 & 65.82 & 38.40 & 17.82 & 5.17 & 1.29 \\
{QCD+UCB} & {Unif} & 95.73 & 113.86 & 87.40 & 50.49 & 23.02 & 8.69 \\
{QCD+UCB} & {Worst} & 125.33 & 89.94 & 66.57 & 31.59 & 9.80 & 3.26 \\
{QCD+klUCB} & {Unif} & 107.36 & 107.98 & 77.02 & 42.57 & 19.51 & 7.68 \\
{QCD+klUCB} & {Worst} & 111.00 & 54.20 & 27.06 & 11.60 & 4.10 & 1.13 \\
\hline
\end{tabular}

\vspace{0.5em} 

\begin{tabular}{c|c|cccccc}
\multicolumn{8}{c}{\textbf{(b) Deterministic Change-Points (Columns: $N_C$)}} \\ \hline
\multicolumn{1}{c|}{\bf Algorithm} & \multicolumn{1}{c|}{\bf PB} & \multicolumn{1}{c}{\bf 2705} & \multicolumn{1}{c}{\bf 875} & \multicolumn{1}{c}{\bf 283} & \multicolumn{1}{c}{\bf 92} & \multicolumn{1}{c}{\bf 30} & \multicolumn{1}{c}{\bf 10} \\
\hline
{MASTER} & {Unif} & 0.00 & 0.00 & 0.00 & 0.00 & 0.00 & 0.00 \\
{MASTER} & {Worst} & 0.00 & 0.00 & 0.00 & 0.00 & 0.00 & 0.00 \\
{GLRklUCB} & {Unif} & 73.83 & 110.55 & 98.46 & 60.01 & 26.89 & 12.25 \\
{GLRklUCB} & {Worst} & 125.41 & 59.89 & 43.19 & 24.44 & 4.53 & 1.58 \\
{QCD+UCB} & {Unif} & 56.00 & 121.00 & 107.00 & 67.00 & 29.00 & 13.00 \\
{QCD+UCB} & {Worst} & 125.00 & 87.00 & 97.00 & 51.00 & 10.00 & 4.00 \\
{QCD+klUCB} & {Unif} & 77.03 & 105.86 & 93.66 & 53.34 & 23.11 & 10.61 \\
{QCD+klUCB} & {Worst} & 116.00 & 48.69 & 24.61 & 13.61 & 4.13 & 1.52 \\
\hline
\end{tabular}

\end{center}
\end{table}

\begin{table}[H]
\caption{Mean Number Of Declared Changes For Different Change-Points, $T=90000$, 4000 Runs.}
\label{table:change_detect_T90000}
\scriptsize
\setlength{\tabcolsep}{3pt} 
\begin{center}
\begin{tabular}{c|c|cccccc}
\multicolumn{8}{c}{\textbf{(a) Random Change-Points (Columns: $\xi$)}} \\ \hline
\multicolumn{1}{c|}{\bf Algorithm} & \multicolumn{1}{c|}{\bf PB} & \multicolumn{1}{c}{\bf 0.3} & \multicolumn{1}{c}{\bf 0.4} & \multicolumn{1}{c}{\bf 0.5} & \multicolumn{1}{c}{\bf 0.6} & \multicolumn{1}{c}{\bf 0.7} & \multicolumn{1}{c}{\bf 0.8} \\
\hline
{MASTER} & {Unif} & 0.00 & 0.00 & 0.00 & 0.00 & 0.00 & 0.00 \\
{MASTER} & {Worst} & 0.00 & 0.00 & 0.00 & 0.00 & 0.00 & 0.00 \\
{GLRklUCB} & {Unif} & 116.07 & 124.83 & 91.32 & 50.66 & 22.68 & 8.66 \\
{GLRklUCB} & {Worst} & 134.99 & 72.26 & 41.53 & 19.21 & 5.68 & 1.37 \\
{QCD+UCB} & {Unif} & 108.69 & 126.67 & 95.84 & 54.55 & 24.21 & 9.00 \\
{QCD+UCB} & {Worst} & 138.30 & 100.33 & 73.03 & 34.17 & 10.77 & 3.56 \\
{QCD+klUCB} & {Unif} & 121.27 & 119.69 & 84.19 & 45.82 & 20.60 & 8.02 \\
{QCD+klUCB} & {Worst} & 121.74 & 59.12 & 29.22 & 12.29 & 4.34 & 1.20 \\
\hline
\end{tabular}

\vspace{0.5em} 

\begin{tabular}{c|c|cccccc}
\multicolumn{8}{c}{\textbf{(b) Deterministic Change-Points (Columns: $N_C$)}} \\ \hline
\multicolumn{1}{c|}{\bf Algorithm} & \multicolumn{1}{c|}{\bf PB} & \multicolumn{1}{c}{\bf 2938} & \multicolumn{1}{c}{\bf 939} & \multicolumn{1}{c}{\bf 300} & \multicolumn{1}{c}{\bf 96} & \multicolumn{1}{c}{\bf 31} & \multicolumn{1}{c}{\bf 10} \\
\hline
{MASTER} & {Unif} & 0.00 & 0.00 & 0.00 & 0.00 & 0.00 & 0.00 \\
{MASTER} & {Worst} & 0.00 & 0.00 & 0.00 & 0.00 & 0.00 & 0.00 \\
{GLRklUCB} & {Unif} & 88.66 & 126.74 & 101.10 & 66.25 & 29.02 & 9.00 \\
{GLRklUCB} & {Worst} & 132.11 & 70.82 & 44.71 & 25.60 & 8.77 & 1.47 \\
{QCD+UCB} & {Unif} & 73.00 & 127.00 & 107.00 & 71.00 & 30.00 & 9.00 \\
{QCD+UCB} & {Worst} & 130.00 & 82.00 & 118.00 & 46.00 & 17.00 & 4.00 \\
{QCD+klUCB} & {Unif} & 93.90 & 123.01 & 93.99 & 59.78 & 24.69 & 8.58 \\
{QCD+klUCB} & {Worst} & 121.04 & 62.76 & 28.57 & 14.45 & 5.01 & 1.15 \\
\hline
\end{tabular}

\end{center}
\end{table}

\begin{table}[H]
\caption{Mean Number Of Declared Changes For Different Change-Points, $T=100000$, 4000 Runs.}
\label{table:change_detect_T100000}
\scriptsize
\setlength{\tabcolsep}{3pt} 
\begin{center}
\begin{tabular}{c|c|cccccc}
\multicolumn{8}{c}{\textbf{(a) Random Change-Points (Columns: $\xi$)}} \\ \hline
\multicolumn{1}{c|}{\bf Algorithm} & \multicolumn{1}{c|}{\bf PB} & \multicolumn{1}{c}{\bf 0.3} & \multicolumn{1}{c}{\bf 0.4} & \multicolumn{1}{c}{\bf 0.5} & \multicolumn{1}{c}{\bf 0.6} & \multicolumn{1}{c}{\bf 0.7} & \multicolumn{1}{c}{\bf 0.8} \\
\hline
{MASTER} & {Unif} & 0.00 & 0.00 & 0.00 & 0.00 & 0.00 & 0.00 \\
{MASTER} & {Worst} & 0.00 & 0.00 & 0.00 & 0.00 & 0.00 & 0.00 \\
{GLRklUCB} & {Unif} & 129.43 & 137.21 & 99.21 & 54.05 & 23.78 & 8.88 \\
{GLRklUCB} & {Worst} & 147.03 & 79.08 & 45.28 & 20.73 & 6.07 & 1.46 \\
{QCD+UCB} & {Unif} & 121.62 & 139.51 & 104.03 & 58.01 & 25.35 & 9.23 \\
{QCD+UCB} & {Worst} & 150.80 & 110.35 & 79.12 & 36.51 & 11.77 & 3.77 \\
{QCD+klUCB} & {Unif} & 134.84 & 131.53 & 91.32 & 48.76 & 21.53 & 8.22 \\
{QCD+klUCB} & {Worst} & 131.94 & 64.20 & 31.33 & 12.93 & 4.59 & 1.27 \\
\hline
\end{tabular}

\vspace{0.5em} 

\begin{tabular}{c|c|cccccc}
\multicolumn{8}{c}{\textbf{(b) Deterministic Change-Points (Columns: $N_C$)}} \\ \hline
\multicolumn{1}{c|}{\bf Algorithm} & \multicolumn{1}{c|}{\bf PB} & \multicolumn{1}{c}{\bf 3163} & \multicolumn{1}{c}{\bf 1000} & \multicolumn{1}{c}{\bf 317} & \multicolumn{1}{c}{\bf 101} & \multicolumn{1}{c}{\bf 32} & \multicolumn{1}{c}{\bf 10} \\
\hline
{MASTER} & {Unif} & 0.00 & 0.00 & 0.00 & 0.00 & 0.00 & 0.00 \\
{MASTER} & {Worst} & 0.00 & 0.00 & 0.00 & 0.00 & 0.00 & 0.00 \\
{GLRklUCB} & {Unif} & 98.45 & 136.94 & 119.82 & 69.49 & 27.38 & 8.31 \\
{GLRklUCB} & {Worst} & 148.35 & 83.78 & 51.12 & 21.52 & 6.74 & 1.19 \\
{QCD+UCB} & {Unif} & 84.00 & 149.00 & 122.00 & 83.00 & 28.00 & 9.00 \\
{QCD+UCB} & {Worst} & 141.00 & 95.00 & 102.00 & 36.00 & 13.00 & 4.00 \\
{QCD+klUCB} & {Unif} & 104.45 & 134.02 & 113.89 & 62.68 & 25.81 & 7.95 \\
{QCD+klUCB} & {Worst} & 130.42 & 71.08 & 34.60 & 14.37 & 4.38 & 1.11 \\
\hline
\end{tabular}

\end{center}
\end{table}

\foreach \x in {1000,2000,5000,10000,20000,30000,40000,50000,60000,70000,80000,90000,100000} {
\begin{figure}[h]
\begin{center}
    \begin{subfigure}[b]{0.49\linewidth}
        \includegraphics[width=\linewidth]{figures_supplement/D_Reg_comp_Normal_\x_4000_suppl.pdf}
        \caption{Geometric Uniform Problem}
        \label{fig:reg_unif\x}
    \end{subfigure}
    \begin{subfigure}[b]{0.49\linewidth}
        \includegraphics[width=\linewidth]{figures_supplement/D_Reg_comp_Normal_Unif_\x_4000_suppl.pdf}
        \caption{Deterministic Uniform Problem}
        \label{fig:reg_det_unif\x}
    \end{subfigure}
    
    \vspace{0.1cm} 
    \begin{subfigure}[b]{0.49\linewidth}
        \includegraphics[width=\linewidth]{figures_supplement/D_Reg_comp_Worst_\x_4000_suppl.pdf}
        \caption{Geometric Worst Problem}
        \label{fig:reg_worst\x}
    \end{subfigure}
    \begin{subfigure}[b]{0.49\linewidth}
        \includegraphics[width=\linewidth]{figures_supplement/D_Reg_comp_Worst_Unif_\x_4000_suppl.pdf}
        \caption{Deterministic Worst Problem}
        \label{fig:reg_det_worst_\x}
    \end{subfigure}

    \vspace{0.5cm} 
    \includegraphics[width=0.52\linewidth]{figures_supplement/Robustness_T_\x_N_4000_suppl.pdf}
    \subcaption{Final dynamic regret versus the decrease in non-stationarity.}
    \label{fig:robust_plot_\x}
    
\end{center}
\caption{Results of the dynamic regret experiments $T=\x$, averaged over $4000$ independent runs.}
\label{fig:results_\x}
\end{figure}
}
\clearpage
\bibliography{main.bib}

\end{document}